\documentclass[10pt,twocolumn,letterpaper]{article}

\usepackage[pagenumbers]{cvpr} 

\usepackage{graphicx}
\usepackage{amsmath}
\usepackage{amssymb}
\usepackage{booktabs}
\usepackage{multirow}
\usepackage{url}
\usepackage{algorithm}
\usepackage{algorithmic}
\usepackage[pagebackref,breaklinks,colorlinks]{hyperref}

\usepackage[capitalize]{cleveref}
\crefname{section}{Sec.}{Secs.}
\Crefname{section}{Section}{Sections}
\Crefname{table}{Table}{Tables}
\crefname{table}{Tab.}{Tabs.}

\def\cvprPaperID{5378} 
\def\confName{CVPR}
\def\confYear{2023}

\begin{document}

\title{

Ultra-low Precision Multiplication-free Training for Deep Neural Networks

}

\newcommand*{\affaddr}[1]{#1} 
\newcommand*{\affmark}[1][*]{\textsuperscript{#1}}
\newcommand*{\email}[1]{\texttt{#1}}
\author{%
Chang Liu\affmark[1,2,3], Rui Zhang\affmark[1,3], Xishan Zhang\affmark[1,3], Yifan Hao\affmark[1,3], Zidong Du\affmark[1], \\ Xing Hu \affmark[1], Ling Li\affmark[2,4], Qi Guo\affmark[1]\\
\affaddr{\affmark[1]SKL of Processor, Institute of Computing Technology, CAS, Beijing, China}\\
\affaddr{\affmark[2]University of Chinese Academy of Sciences, Beijing, China}\\
\affaddr{\affmark[3]Cambricon Technologies, Beijing, China}\\
\affaddr{\affmark[4]SKL of Computer Science, Institute of Software, CAS}\\
\email{\{liuchang18s, zhangrui, zhangxishan, }\\
\email{ haoyifan, duzidong, huxing, guoqi\}@ict.ac.cn}\\
\email{\{liling\}@iscas.ac.cn}\\
}
\maketitle
\begin{abstract}

The training for deep neural networks (DNNs) demands immense energy consumption, which restricts the development of deep learning as well as increases carbon emissions. Thus, the study of energy-efficient training for DNNs is essential. In training, the linear layers consume the most energy because of the intense use of energy-consuming full-precision (FP32) multiplication in multiply–accumulate (MAC). The energy-efficient works try to decrease the precision of multiplication or replace the multiplication with energy-efficient operations such as addition or bitwise shift, to reduce the energy consumption of FP32 multiplications. However, the existing energy-efficient works cannot replace all of the FP32 multiplications during both forward and backward propagation with low-precision energy-efficient operations. In this work, we propose an Adaptive Layer-wise Scaling PoT Quantization (ALS-POTQ) method and a Multiplication-Free MAC (MF-MAC) to replace all of the FP32 multiplications with the INT4 additions and 1-bit XOR operations. In addition, we propose Weight Bias Correction and Parameterized Ratio Clipping techniques for stable training and improving accuracy. In our training scheme, all of the above methods do not introduce extra multiplications, so we reduce up to 95.8\% of the energy consumption in linear layers during training. Experimentally, we achieve an accuracy degradation of less than 1\% for CNN models on ImageNet and Transformer model on the WMT En-De task. In summary, we significantly outperform the existing methods for both energy efficiency and accuracy. 

\end{abstract}

\section{Introduction}
\label{sec:intro}
\begin{figure}[t]
  \centering
  \includegraphics[scale=0.56]{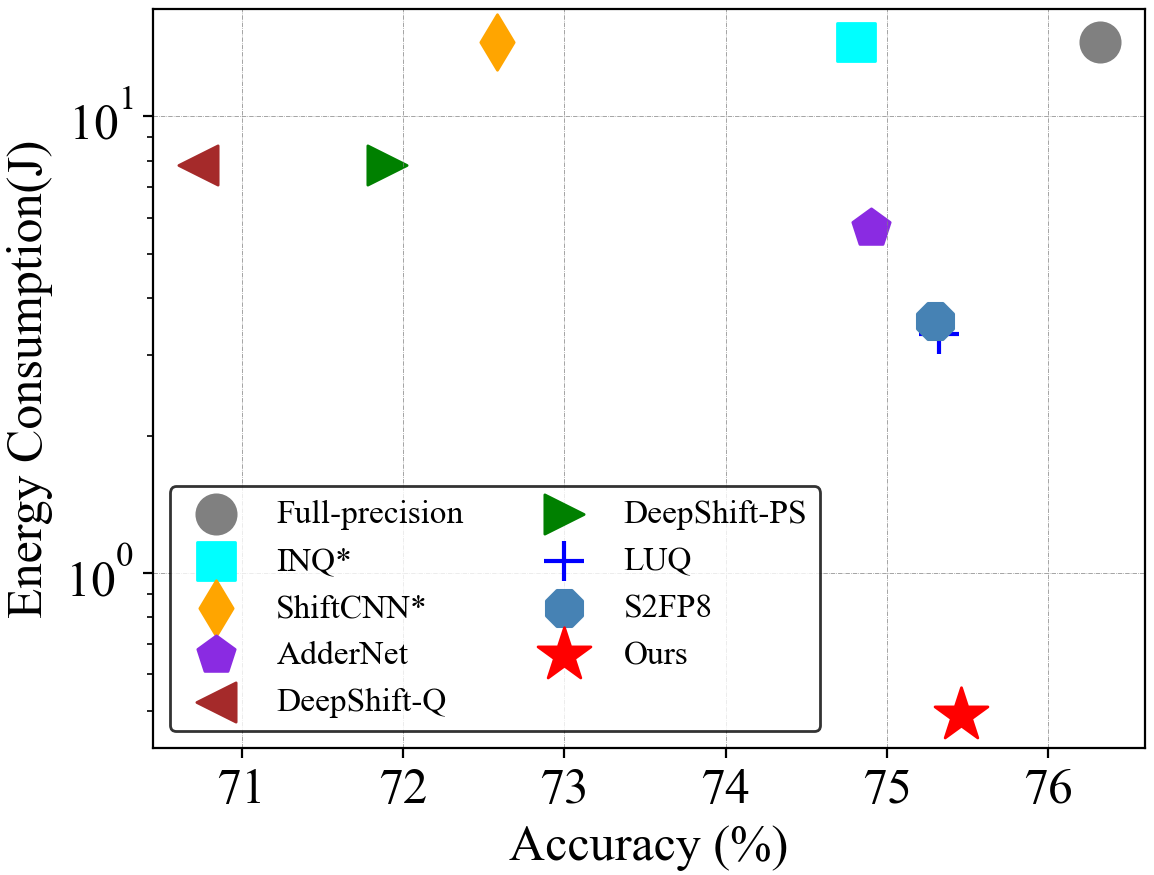}
  \caption{Energy-Accuracy joint comparison. ``Accuracy'' refers to the accuracy results of training ResNet50 on ImageNet from scratch. ``Energy Consumption'' refers to the energy consumption of MACs for training ResNet50 on ImageNet at one iteration. Note that INQ and ShiftCNN apply their method to the pre-train models, so their training consumption is the same as full-precision training.}
  \label{fig:pareto}
\end{figure}
In recent years, deep neural networks (DNNs) have achieved remarkable success in many AI applications. However, this success comes at the cost of training the models, which consumes substantial energy. For example, training a BERT$_{base}$ model demands energy consumption of 948 $kw\cdot  h$, which is more than the global average household electricity consumption per capita per year (731 $kw\cdot h$) ~\cite{strubell2019energy,electricity}.
Furthermore, the immense energy consumption of DNN training significantly increases carbon emissions, leading to a negative impact on the global climate~\cite{anthony2020carbontracker,thompson2020computational,aicompute}. Thus, reducing the energy consumption of training for DNNs is desperately required. 

In DNN training, the linear layers (including convolutional and fully-connected layers) consume the most energy because of the intense use of FP32 energy-consuming multiplication in multiply–accumulate (MAC). 
The FP32 multiplication is energy-consuming partly because of its high precision.
For example, the energy consumption of an FP32 multiplication is approximately 4x higher than that of FP16 multiplication.
Thus, to obtain energy-efficient DNNs, there are many quantization methods replacing the full-precision multiplications with low-precision multiplications by quantization techniques. Some quantization methods start with the full-precision (FP32) pre-trained model~\cite{2020Post,ron2019post}, hence they cannot reduce the energy consumption of training. The other methods train models from scratch by quantizing the weights ($W$), activations ($A$), and activation gradients ($G$) to 16-bit~\cite{das2018mixed, koster2017flexpoint:} or 8-bit~\cite{wang2018training,zhu2019towards,sun2019hybrid,cambier2020shifted}.  

Besides, the multiplication itself has significantly higher energy consumption than energy-efficient operations such as addition and bitwise shift. For example, the energy consumption of an INT32 multiplication is approximately 22x higher than that of an INT32 addition. Thus, there are multiplication-less methods that directly replace the multiplication with energy-efficient operations such as addition and bitwise shift
~\cite{zhou2017incremental,miyashita2016convolutional,gudovskiy2017shiftcnn,elhoushi2021deepshift, chen2020addernet, you2020shiftaddnet, chmiel2021logarithmic}.
Most of these works, such as INQ~\cite{zhou2017incremental}, ShiftCNN~\cite{gudovskiy2017shiftcnn}, and LogNN~\cite{miyashita2016convolutional} also start with the FP32 pre-trained models rather than training from scratch so they cannot reduce the energy consumption of training.
Among the methods that can train from scratch~\cite{chen2020addernet,elhoushi2021deepshift, chmiel2021logarithmic}, AdderNet~\cite{chen2020addernet} replaces all of the FP32 multiplications in the linear layer with FP32 additions whose energy consumption is still higher than the fixed point operations. 
The other works~\cite{elhoushi2021deepshift, chmiel2021logarithmic} apply low-precision Power-of-Two (PoT) numbers, whose value is zero or power of 2, to replace a part of the multiplications in training with bitwise shifts and sign flip operations.
However, they cannot replace all of the multiplications during forward or backward propagation. For example, DeepShift~\cite{elhoushi2021deepshift} only converts $W$ to 5-bit PoT numbers because only the value range of $W$ can be limited to the representation range of their 5-bit PoT numbers. Similarly, LUQ~\cite{chmiel2021logarithmic} only converts $G$ to PoT numbers because their method cannot approximate distributions of $W$ and $A$ well. These methods cannot use one data format to represent each of $W$, $A$, and $G$ whose data ranges and distributions vary widely from each other, so they keep one-third of the multiplications in training.

In summary, the existing quantization methods and multiplication-less methods cannot replace all of the FP32 multiplications during both forward and backward propagation with low-precision energy-efficient operations. Thus the training energy consumption savings of all these works are limited, as shown in Figure~\ref{fig:pareto}.

In this work, we propose a Multiplication-Free MAC (MF-MAC) to replace the FP32 multiplication with an INT4 addition and a 1-bit XOR operation, which are significantly more energy-efficient than the existing methods. To support the MF-MAC, we propose an Adaptive Layer-wise Scaling PoT Quantization (ALS-PoTQ) method that accommodates all the data range of $W$, $A$, and $G$ to convert them to unified 5-bit PoT numbers. It is also important to note that the proposed ALS-PoTQ method does not introduce extra multiplications while the exiting methods~\cite{wang2018training,zhu2019towards,sun2019hybrid,cambier2020shifted,chmiel2021logarithmic} introduce multiplications when quantizing data. Thus, all of the operations in the proposed MF-MAC and ALS-PoTQ are energy-efficient.

In addition, to keep training stable and improve accuracy, we propose a Weight Bias Correction (WBC) technique to correct the bias of $W$, and a Parameterized Ratio Clipping (PRC) technique to avoid the rigid resolution problem of $A$. These two techniques also do not introduce extra multiplications. Finally, by combing all of the above techniques, the complete multiplication-free training scheme replaces all of the multiplications during forward and backward propagation with unified energy-efficient operations, which the existing methods cannot achieve.

Applying our multiplication-free training scheme, we conduct experiments for AlexNet~\cite{krizhevsky2012imagenet}, ResNet18~\cite{he2016deep}, ResNet50~\cite{he2016deep} models on ImageNet~\cite{deng2009imagenet:} and Transformer model on the WMT En-De task, achieving an accuracy degradation of less than 1\%. 
Compared with FP32 training, we reduce up to 95.8\% of the energy consumption in linear layers during training with negligible accuracy degradation.
To sum up, our method significantly outperforms the existing methods for both energy efficiency and accuracy, as shown in Figure~\ref{fig:pareto}.
To the best of our knowledge, it is the first work to replace all of the multiplications with low-precision additions during both forward and backward propagation in DNN training.

Our contributions can be listed as follows:
\begin{itemize}
    \item  We propose a Multiplication-free MAC (MF-MAC) with an Adaptive Layer-wise Scaling PoT Quantization (ALS-POTQ) method to replace the FP32 multiplication in MAC with an INT4 addition and an XOR operation.
    \item We propose two multiplication-free techniques, Weight Bias Correction (WBC) and Parameterized Ratio Clipping (PRC), that can keep training stable and improve accuracy.
    \item  Our method reduces up to 95.8\% of the energy consumption in linear layers during training compared with FP32 training with an accuracy degradation of less than 1\%. It significantly outperforms other methods for both energy -efficiency and accuracy results.
\end{itemize}

\begin{figure*}[t]
  \centering
  \begin{subfigure}{0.33\linewidth}
    \includegraphics[scale=0.35]{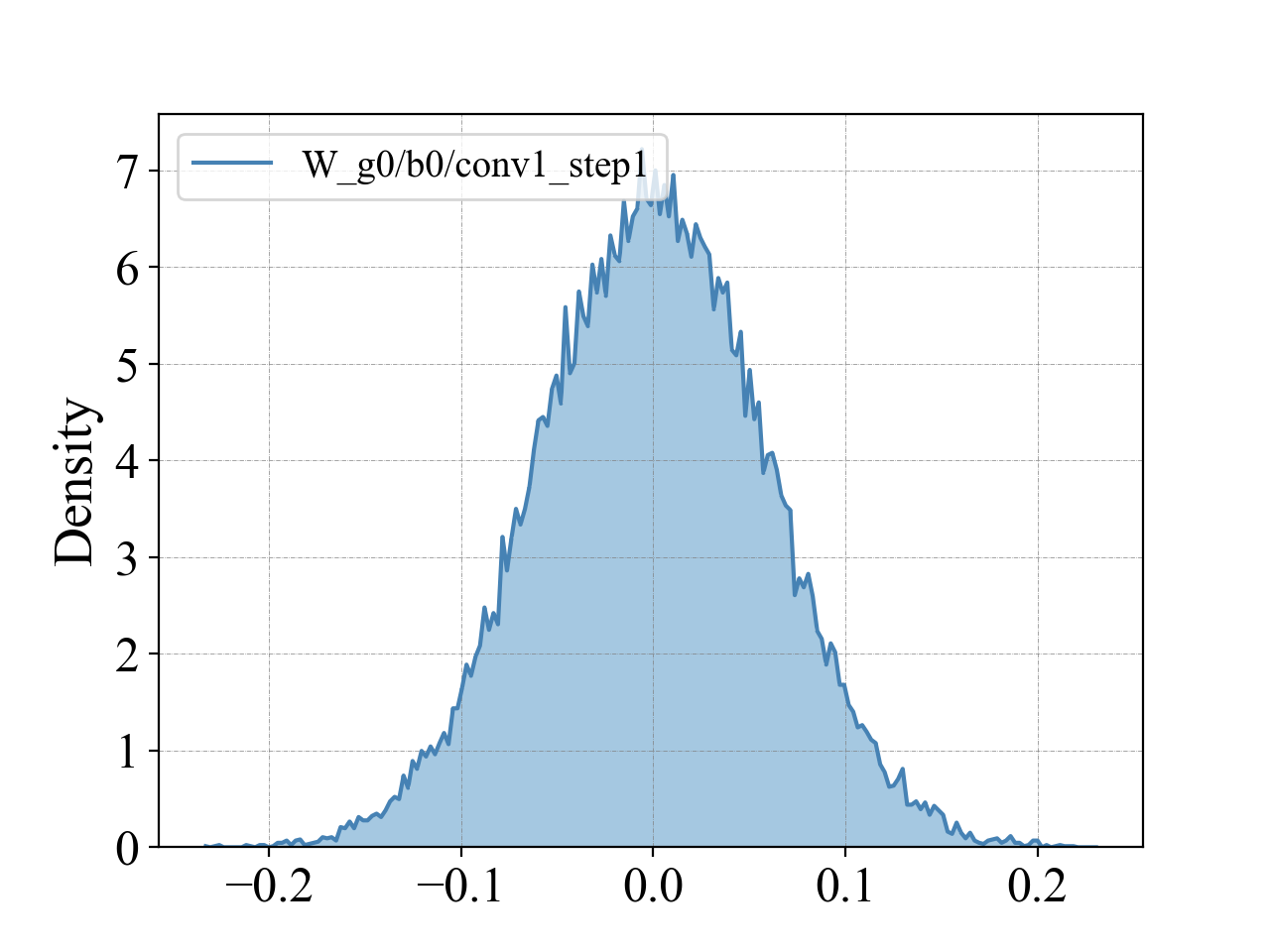}
    \caption{Weight Distribution.}
    \label{fig:dist-w}
  \end{subfigure}
  \hfill
    \begin{subfigure}{0.33\linewidth}
    \includegraphics[scale=0.35]{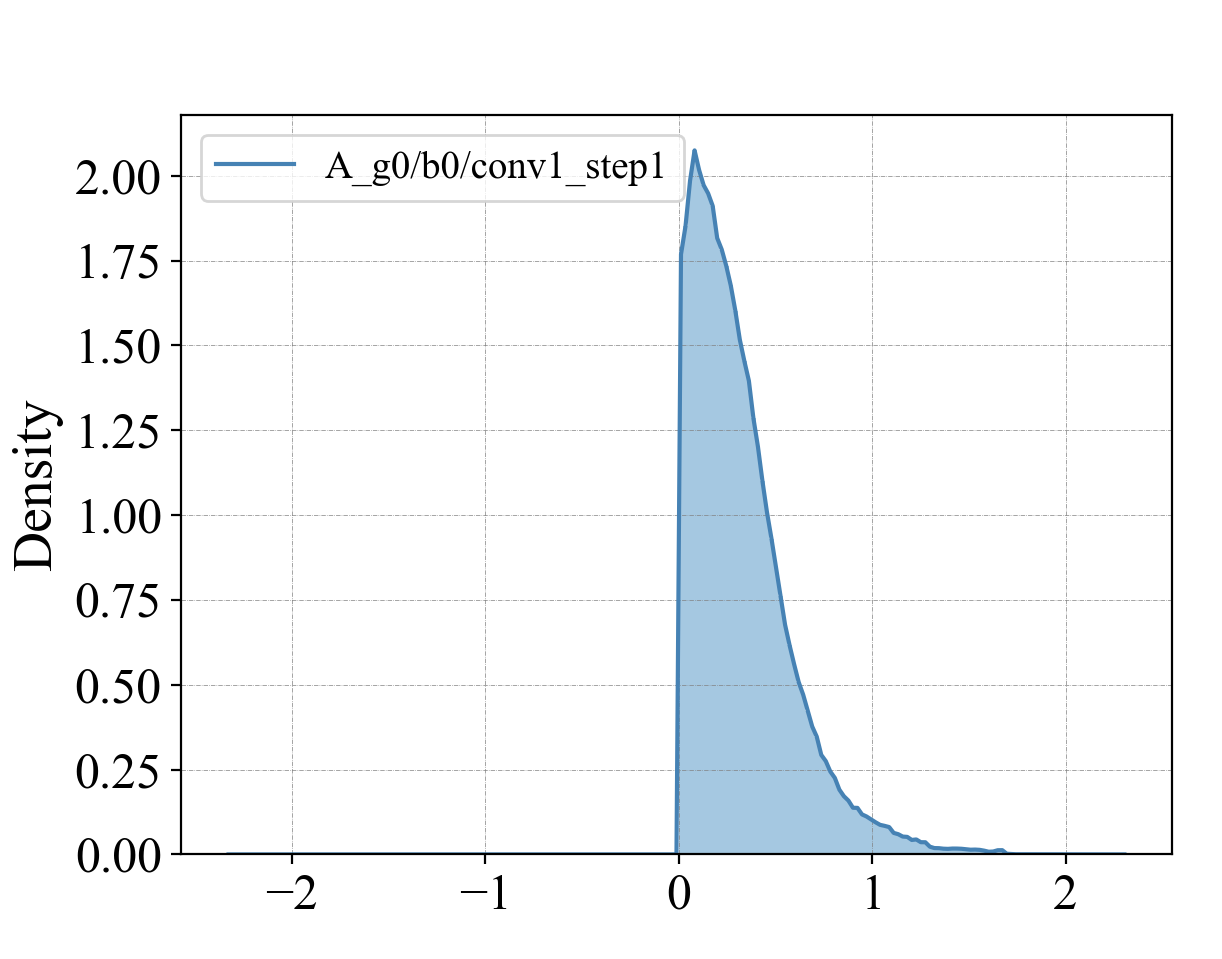}
    \caption{Activation Distribution.}
    \label{fig:dist-a}
  \end{subfigure}
  \hfill
  \begin{subfigure}{0.33\linewidth}
    \includegraphics[scale=0.35]{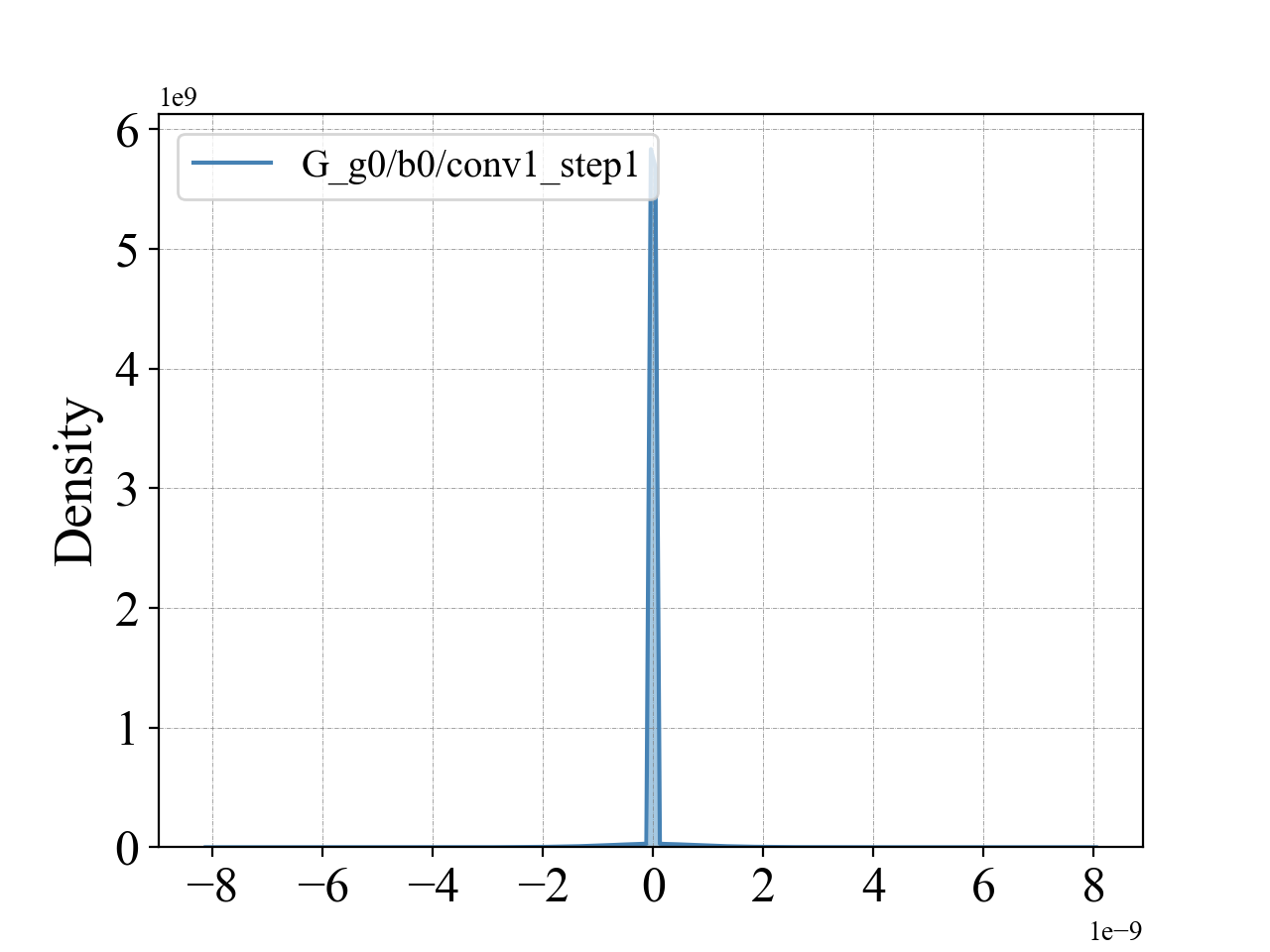}
    \caption{Gradient distribution.}
    \label{fig:dist-g}
  \end{subfigure}
  \begin{subfigure}{0.33\linewidth}
    \includegraphics[scale=0.35]{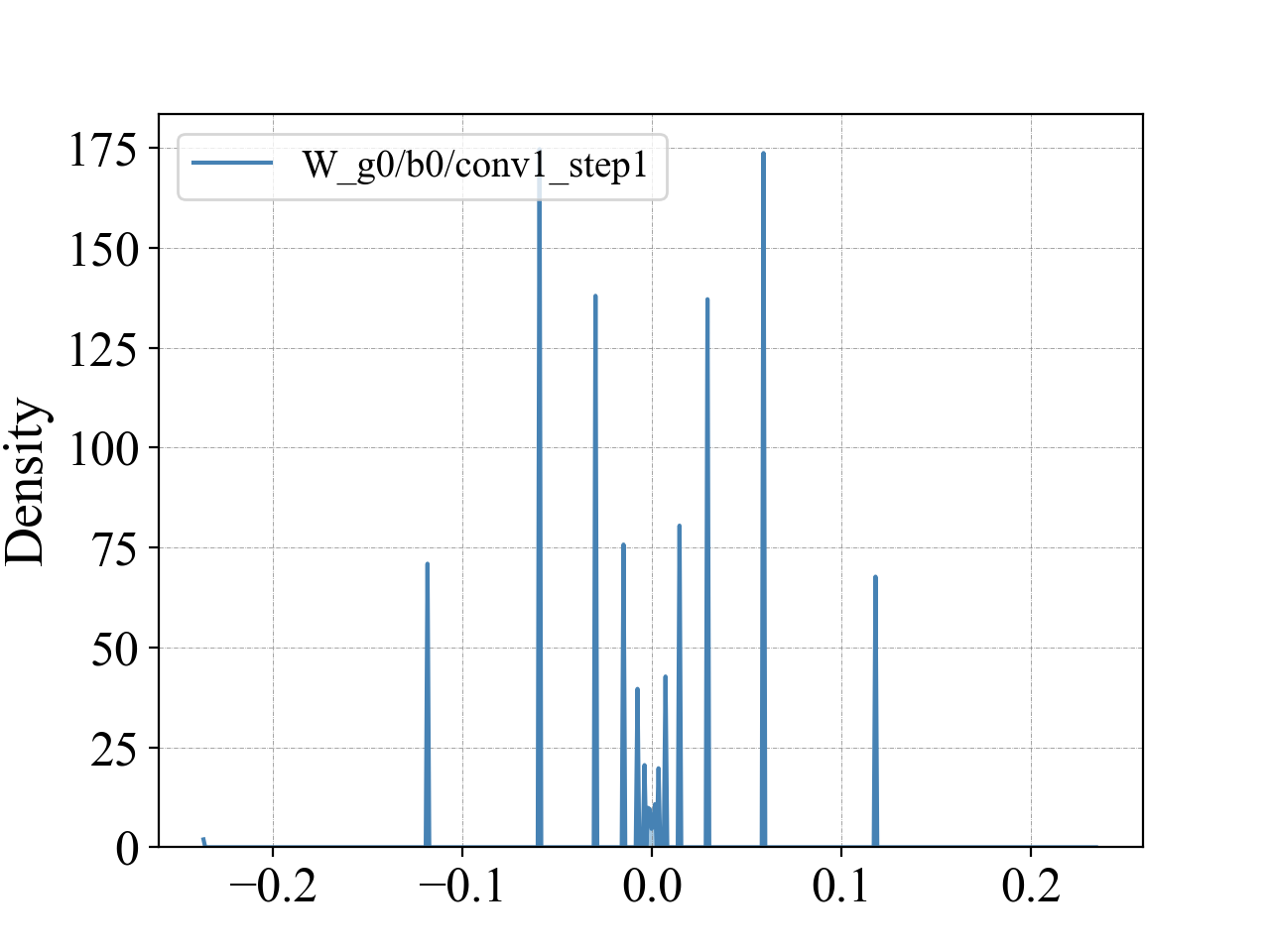}
    \caption{PoT Quantized Weight Distribution.}
    \label{fig:dist-w-q}
  \end{subfigure}
  \hfill
    \begin{subfigure}{0.33\linewidth}
    \includegraphics[scale=0.35]{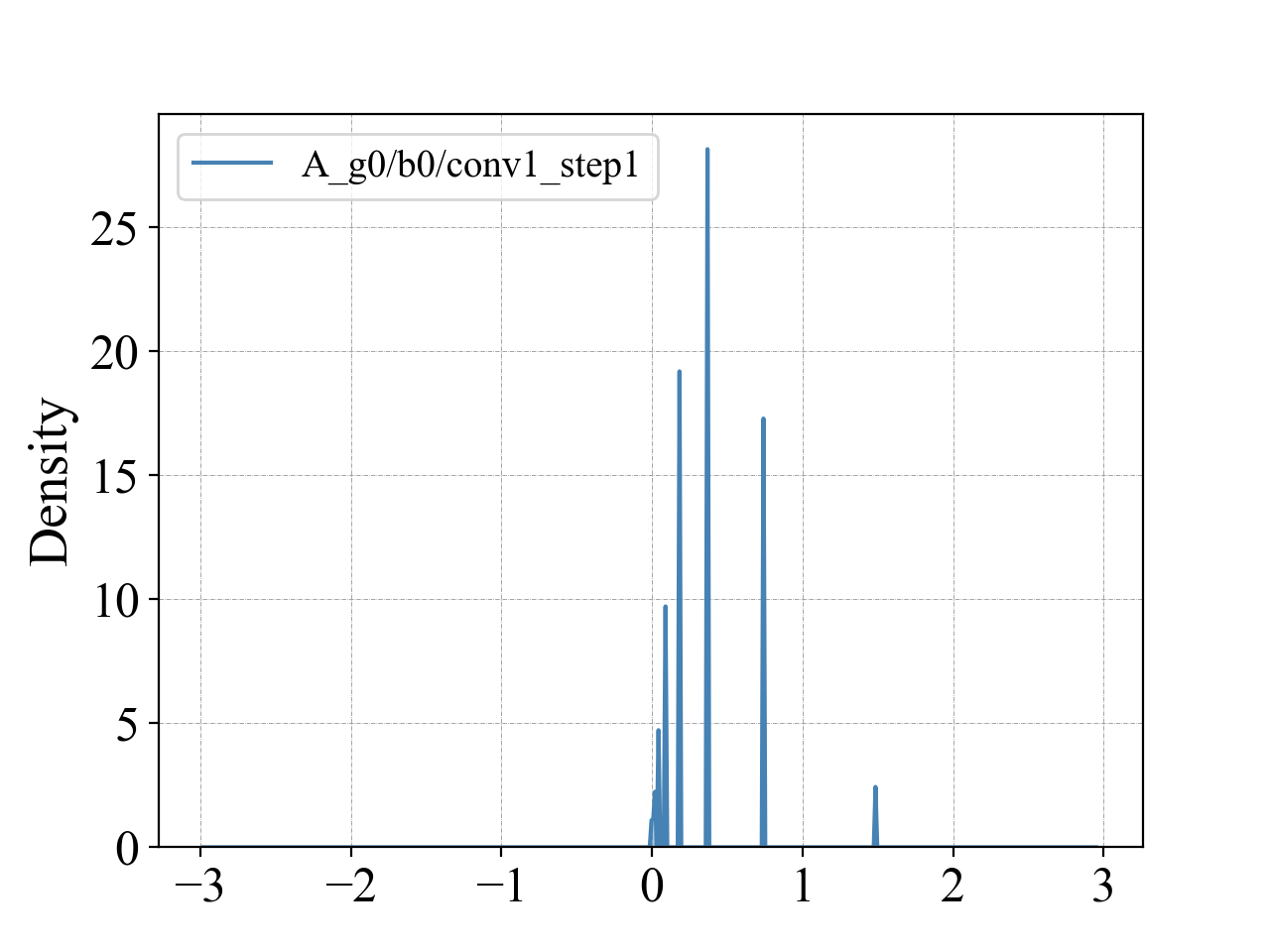}
    \caption{PoT Quantized Activation Distribution.}
    \label{fig:dist-a-q}
  \end{subfigure}
  \hfill
  \begin{subfigure}{0.33\linewidth}
    \includegraphics[scale=0.35]{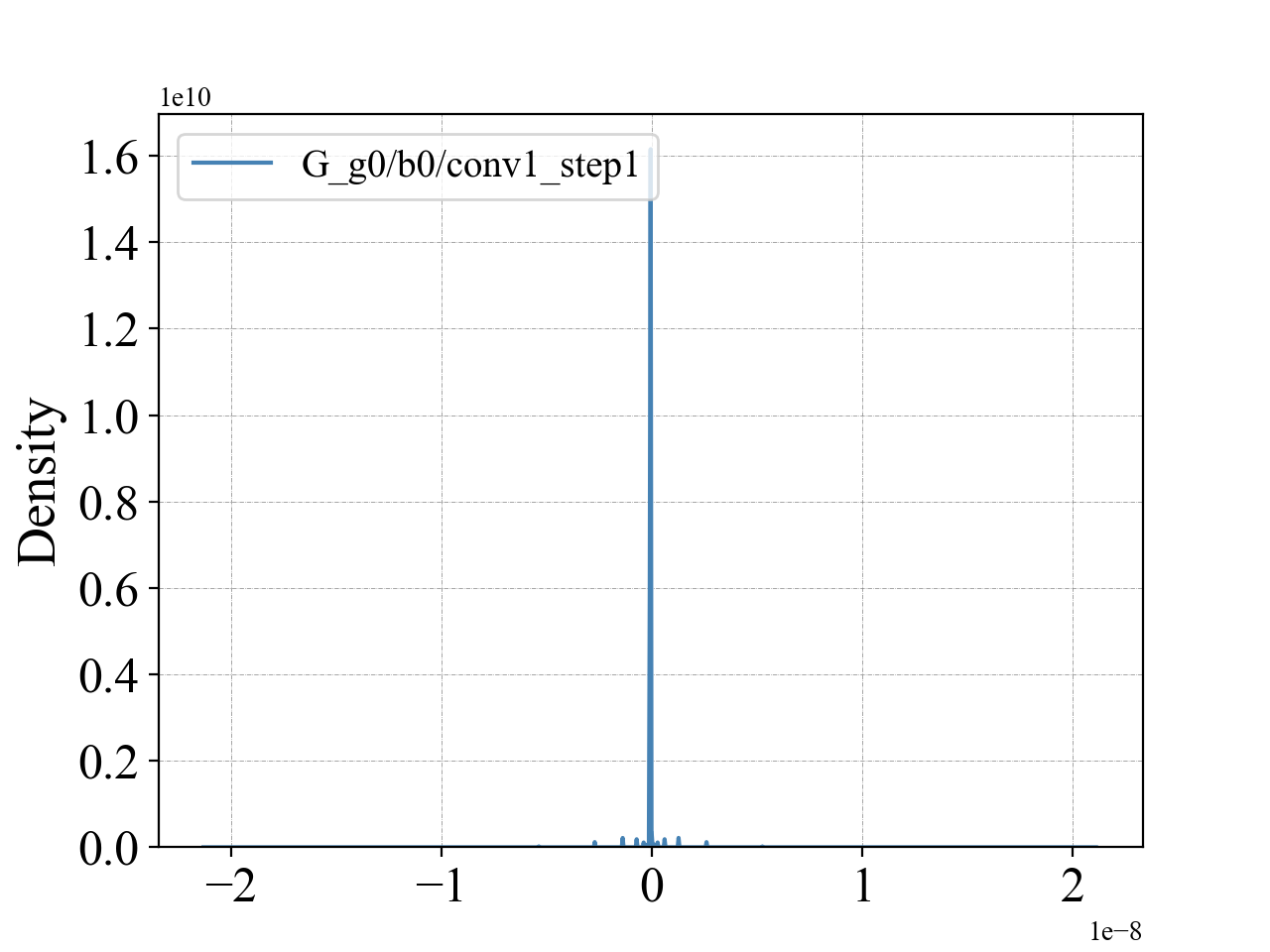}
    \caption{PoT Quantized Gradient distribution.}
    \label{fig:dist-g-q}
  \end{subfigure}
  \caption{Distributions of W, A, and G and their ALS-PoTQ results for ResNet50 on ImageNet. }
  \label{fig:distribution}
\end{figure*}

\section{Related Work}
\label{sec:related_work}

Recently, to avoid the high energy-consuming of the multiplications in deep learning, many low-precision works quantize the $W$, $A$, or $G$ to low-precision before multiplication~\cite{zhou2016dorefa-net:,das2018mixed, koster2017flexpoint:,wang2018training,zhu2019towards,ron2019post,2021FAT,2020Bayesian,2020Rethinking,2020Least,2020Position,qin2020binary,courbariaux2015binaryconnect:,hubara2016binarized,sun2019hybrid,cambier2020shifted}. Among these works, some quantization methods start with the full-precision (FP32) pre-trained model~\cite{2020Post,ron2019post}, hence they cannot reduce the energy consumption of training. The other methods train models from scratch by quantizing the weights $W$, activations $A$, and activation gradients $G$ to 16-bit numbrs~\cite{das2018mixed, koster2017flexpoint:}, 8-bit numbers~\cite{wang2018training,zhu2019towards,sun2019hybrid,cambier2020shifted}, or other formats such as radix-4 numbers~\cite{sun2020ultra}.

Besides, there are some multiplication-less works that directly replace multiplication with energy-efficient operations such as additions~\cite{chen2020addernet}, bitwise shifts~\cite{zhou2017incremental,gudovskiy2017shiftcnn,miyashita2016convolutional,elhoushi2021deepshift} or the combination of them~\cite{you2020shiftaddnet}. AdderNet~\cite{chen2020addernet} takes the $l$1-norm distance between filters and input feature as
the output response and replaces the multiplications in the linear layers with additions. The network is still computed and stored with the FP32 numbers. This work achieves approximately 3\% of accuracy degradation for the ResNet models on ImageNet.
The methods replacing multiplications with bitwise shifts are based on a logarithm quantization method. The logarithm quantization method quantizes full precision data to zeros and powers of a radix number. For example, Ultra-Low training~\cite{sun2020ultra} uses radix-4 logarithm format to represent gradients, which need specialized hardware support. When the radix of the logarithmic function is 2, it is called power-of-two quantization and the multiplications can be replaced with a bitwise shift.
Incremental Network Quantization (INQ)~\cite{zhou2017incremental} partitions the pre-trained weights into two sets, one of which is PoT quantized while the other is retrained to compensate for accuracy degradation.
ShiftCNN~\cite{gudovskiy2017shiftcnn} quantizes $W$ of a pre-trained model to PoT representation. 
LogNN~\cite{miyashita2016convolutional} quantizes pre-trained $W$ and $A$ to 4-bit PoT numbers. All of the above methods are applied to the pre-trained FP32 models rather than training from scratch. Hence, they cannot be used to reduce the energy consumption of training. Besides, some works can reduce the energy consumption for training: Deepshift~\cite{elhoushi2021deepshift} converts all $W$ to PoT numbers with two training methods DeepShift-Q and DeepShift-PS, achieving an accuracy degradation of 4.42\%$\sim$5.59\%. Logarithmic Unbiased Quantization(LUQ)~\cite{chmiel2021logarithmic} applies a logarithmic unbiased quantization with pruning operation to quantization gradients during training, achieving an accuracy degradation of less than 2\%. These works replace a part of the multiplications in training with bitwise shifts. However, they keep at least one-third of the multiplications during forward or backward propagation.

In summary, the existing works cannot replace all of the multiplications with the most energy-efficient low-precision fixed-point additions, during both forward and backward propagation.

\section{Preliminaries}

\label{sec:pot-quant}
In this section, we give a definition of Power-of-Two (PoT) quantization and how the multiplication of numbers after PoT quantization is calculated.
The value of a Power-of-Two number $p$ is either power of two or zero:
\begin{equation}
\label{eq:pot}
   \{0, \pm 2^{-2^{b-2}+1}, \pm 2^{-2^{b-2}+2}, \cdots, \pm 2^{2^{b-2}-1}\},
\end{equation}
where $b$ is the bit-width to represent the number. The bit-width $b$ contains 1 sign bit and $b-1$ exponent bits.

To convert a FP32 number $f$ to a $b$-bit PoT number $p$, a basic PoT quantization method is as follows:
\begin{equation}
    e = Round(log_2(\vert f\vert))
\end{equation}
\begin{equation}
\label{quant}
p =\left\{
\begin{array}{lc}
0, \qquad \qquad \qquad \qquad if \quad  e < -2^{b-2}+1,  \\ 
sign(f)\cdot 2^{2^{b-2}-1}, \quad \quad if \quad e \geq 2^{b-2}-1, \\
sign(f) \cdot 2^e, \quad \quad \qquad else, 
\end{array}
\right.
\end{equation}
where $e$ is the exponent of a PoT number $p=2^e$, $round$ refers to round-to-nearest.

After converting the FP32 numbers to $b$-bit PoT numbers, the multiplication between two $b$-bit PoT numbers $2^k$ and $2^m$ can be replaced with a $(b-1)$-bit fixed-point addition in the logarithm domain and a 1-bit sign flip operation:
\begin{equation}
    2^{k}\cdot 2^{m} = 2^{k+m},
\end{equation}
where $k,m \in [-2^{b-2}+1, 2^{b-2}-1]$. The 1-bit sign flip operation can be formulated as, 
\begin{equation}
flip(s_1, s_2)= s_1 \oplus s_2,
\label{eq:sign}
\end{equation}
where $s_1$ and $s_2$ are the sign bits of the two numbers. If a PoT number's sign bit is 1, the PoT number is negative. If the sign bit is 0, the PoT number is positive.

In addition, the multiplication between a $b$-bit PoT number $s\cdot2^k$ and a fixed-point number $x$ can be replaced with a bitwise shift and a 1-bit sign flip. The bitwise shift can be formulated as,  
\begin{equation}
2^{k}\cdot x =\left\{
\begin{array}{lc}
x \ll k, \qquad if \quad k>0,  \\ 
x \gg k, \qquad if \quad k<0, \\
x, \quad \qquad if \quad k==0.
\end{array}
\right.
\end{equation}


\begin{figure}[t]
  \centering
  \includegraphics[scale=0.28]{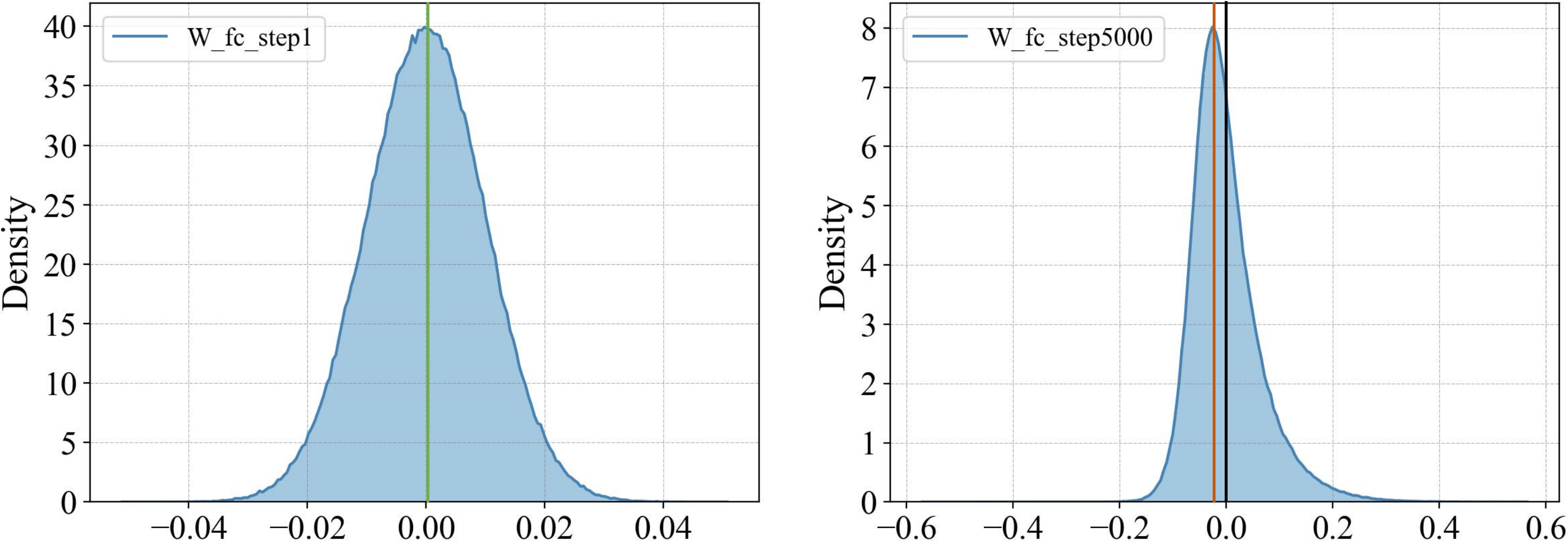}
  \caption{Weight Distributions (Density) at different steps. The green and orange lines refers to the mean value of weights at step 1. The orange line refers to the mean value of weights at step 5000.}
  \label{fig:weight}
\end{figure}

\section{Methodology}
\label{sec:method}

To replace all the multiplications with energy-efficient operations, we aim to convert all of the FP32 $W$, $A$, and $G$ in the DNN training to PoT numbers with negligible accuracy degradation. 

\begin{figure}
  \centering
  \begin{subfigure}{0.48\linewidth}
    \includegraphics[scale=0.275]{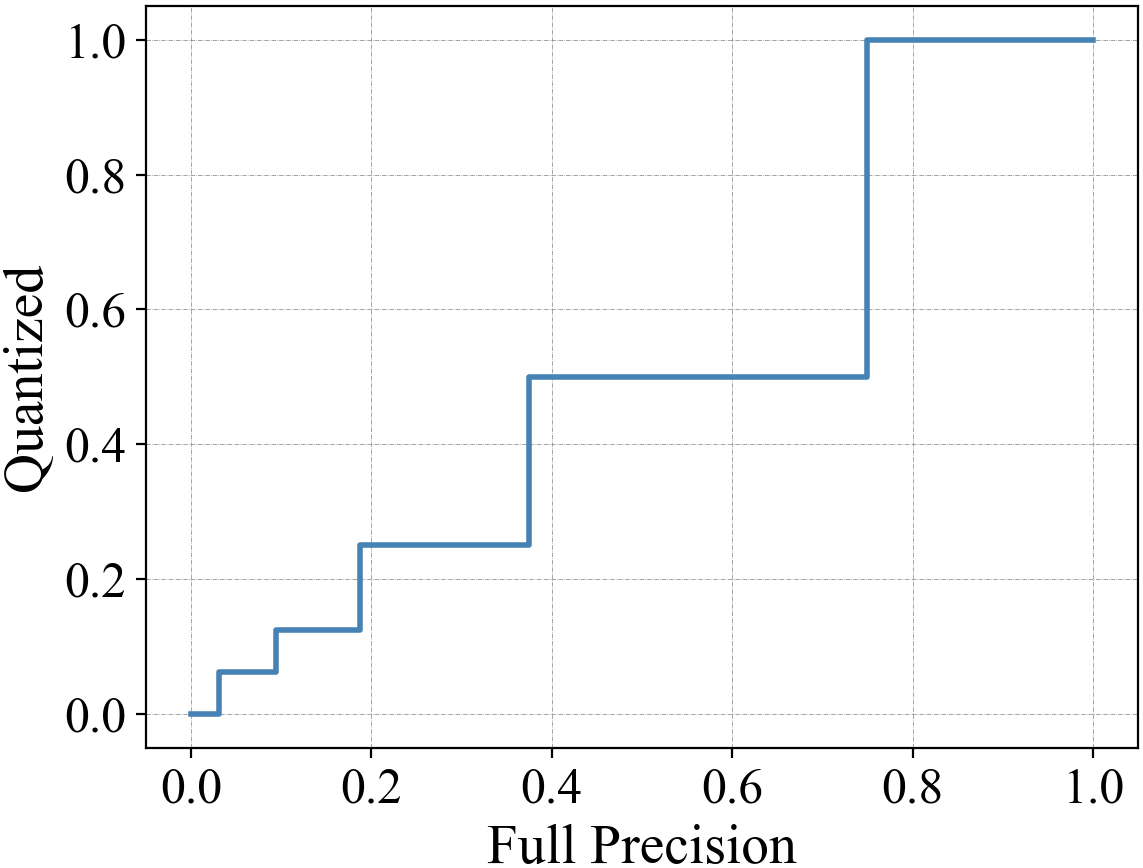}
    \caption{3-bit PoT Quantization.}
    \label{fig:pot3}
  \end{subfigure}
  \hfill
    \begin{subfigure}{0.48\linewidth}
    \includegraphics[scale=0.275]{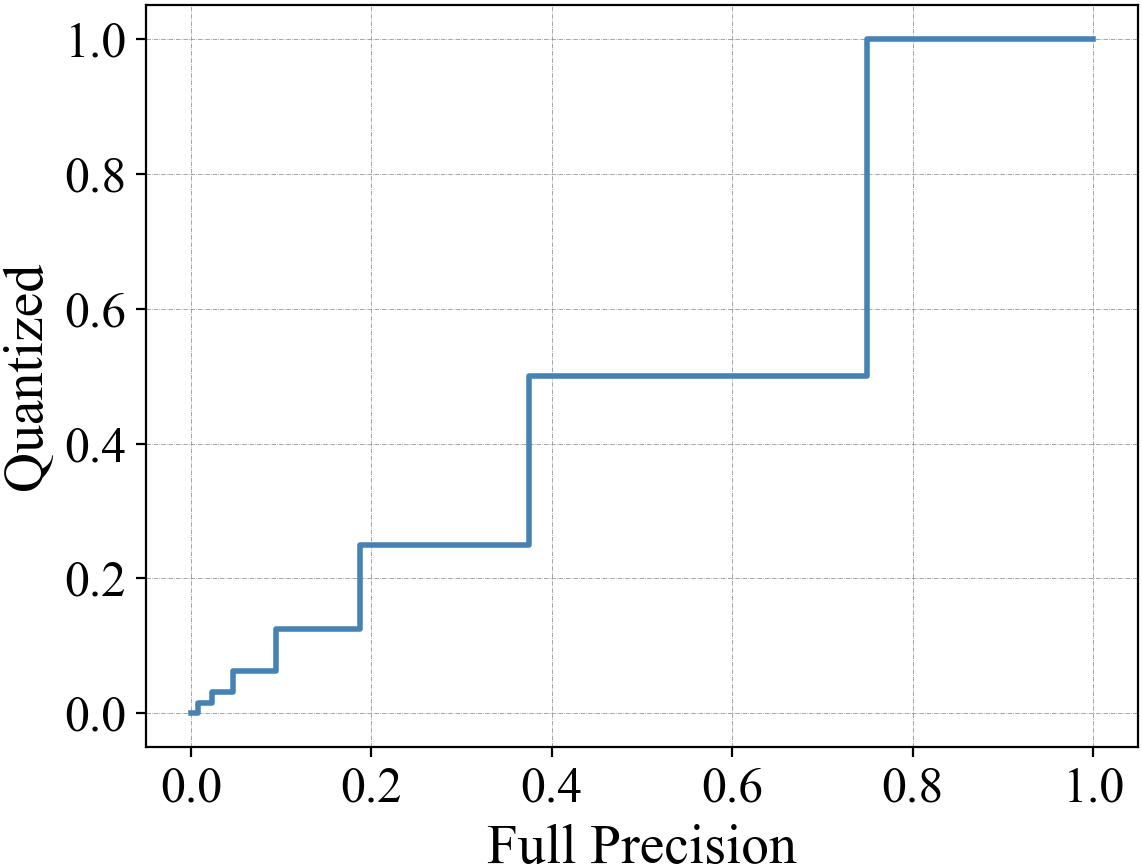}
    \caption{4-bit PoT Quantization.}
    \label{fig:pot4}
  \end{subfigure}
  \caption{PoT quantization to 3-bit and 4-bit PoT number. For clearer illustration, the full-precision data is normalized.}
  \label{fig:pot}
\end{figure}

\begin{figure*}[t]
  \centering
  \includegraphics[scale=0.3]{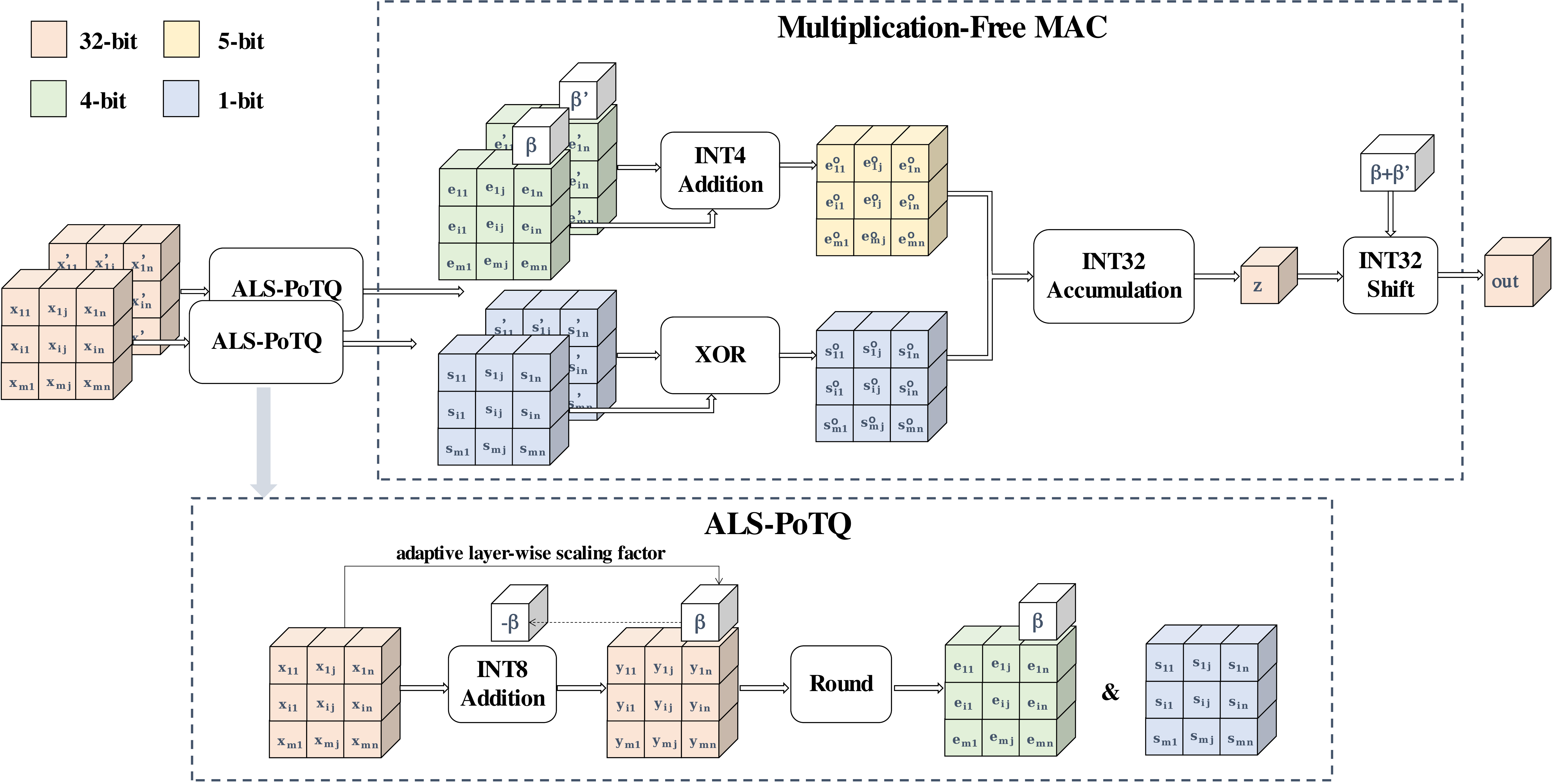}
  \caption{Implementation of Multiplication-Free MAC with ALS-PoTQ. The orange data is 32-bit, the yellow data is 5-bit, the green data is 4-bit, and the blue data is 1-bit.}
  \label{fig:compute_mac}
\end{figure*}
\subsection{Adaptive Layer-wise Scaling PoT Quantization}
\label{sec:als-pot}

We first observe their distributions in extensive layers and networks at different training steps. We find that the distributions of $W$, $A$, and $G$ in DNNs are all spiky and long-tailed near-lognormal distributions, as shown in Figure~\ref{fig:distribution}\subref{fig:dist-a}\subref{fig:dist-w}\subref{fig:dist-g}. (For more distribution figures, please refer to Appendix A.) In other words, most of the numbers concentrate around zero (peak area) and a few numbers are of relatively high magnitude. Consistently, the resolution of PoT numbers is dense in the region near zero and sparse in the regions far from zero. This consistency provides the foundation for applying PoT quantization to DNNs.

However, the basic PoT quantization method cannot be applied to different $W$, $A$, and $G$ directly. The representation range of $b$-bit PoT numbers is $[-2^{2^{b-2}-1}, 2^{2^{b-2}-1}]$, while the data ranges of $W$, $A$, and $G$ are different and change along with the layer and training step as shown in Figure~\ref{fig:distribution}. For a practical hardware implementation, the bit-width should be fixed. Thus, to accommodate the representation range of fixed bit-width PoT numbers, we apply an adaptive scaling factor $\alpha$ to scale the $W$, $A$, and $G$ before the PoT quantization.

Suppose a set of FP32 (i.e., 32-bit floating-point) number $F = \{f_1, f_2, \cdots, f_n\}$. ($F$ can be the $W$, $A$, or $G$ in a linear layer.) We limit the data range of $F$ to the $b$-bit representation range of PoT format, \ie,  $[-2^{2^{b-2}-1}, 2^{2^{b-2}-1}]$. The layer-wise scaling factor $\alpha$ is:
\begin{equation}
\label{eq:alpha}
    \alpha = \frac{max(\vert F \vert)}{2^{2^{b-2}-1}}
\end{equation}
\begin{equation}
\label{eq:scaling}
    F_{scaled} = \frac{F}{\alpha} 
\end{equation}
Then, we quantize $F_{scaled}$ to the PoT numbers $P$ as described in Section~\ref{sec:pot-quant}. 
Thus, the real value of quantized $\hat{F}$ is 
\begin{equation}
    \hat{F} = \alpha \cdot P.
\end{equation}

The scaling operation introduces extra multiplications as shown in Equation~\eqref{eq:scaling}, which is contrary to our purpose. Thus, we further round $\alpha$ to the nearest PoT number:
\begin{equation}
    \beta = Round(log_2(\alpha)),
\end{equation}
where $\beta$ is an integer and different for $W$, $A$, and $G$ in various layers and step. Its value is approximately in the range $[-5,-2]$ for $W$ and $A$, and $[-20,-10]$ for $G$ empirically.

 Then, the multiplications in Equation~\eqref{eq:scaling} can be replaced with the additions between $\beta$ with the exponent part of $F$.  Moreover, the consumption of $\alpha$'s storage and the scalar multiplication in Equation~\eqref{eq:alpha} can be ignored because $\alpha$ is layer-wise. There is only one $\alpha$ for tens of thousands of $W$ (or $A$, $G$) in a layer.

In summary, we can convert each $W$, $A$, and $G$ to $b$-bit PoT numbers along with the Adaptive Layer-wise Scaling PoT Quantization (ALS-PoTQ) method without introducing extra multiplications. In this work, we choose $b=5$, so each FP32 multiplication in MAC can be replaced with an INT4 addition and a 1-bit sign flipping. The distributions of $W$, $A$, and $G$, and corresponding ALS-PoTQ quantized data are shown in Figure~\ref{fig:distribution}. We can observe that although the distributions of $W$, $A$, and $G$ vary widely, the ALS-PoTQ quantized data can fit each of them well.

\subsection{Weight Bias Correction}
\label{sec:weight_correct}

In practical experiments, we find that the distribution of W changes frequently and its mean deviates during training as shown in Figure~\ref{fig:weight}. The weights with biases are not consistent with the symmetry of PoT quantization as shown in Figure~\ref{fig:distribution}\subref{fig:dist-w-q}. Thus, the mean-square-error between quantized $\hat{W}$ and original $W$ becomes larger. In addition, W's bias can be accumulated to the activation gradients ($G$) in backward propagation, and the bias on $G$ is theoretically and empirically proven to impact the training convergence ~\cite{liu2022rethinking}. Thus, we propose a weight bias correction technique that is effective to eliminate the bias. 

Given a set of FP32 weights $W=\{w_1, w_2, \cdots, w_n\}$, we use the weight bias correction technique to obtain unbiased weights $\tilde{W}$:

\begin{equation}
\tilde{W} = W -mean(W)
\end{equation}

We apply the Weight Bias Correction (WBC) technique to $W$ before ALS-PoTQ. With this technique, the training converges normally. Unlike the weight normalization technique used in previous work~\cite{li2019additive}, the weight bias correction technique does not introduce any extra multiplication.

\subsection{Parameterized Ratio Clipping}

When applying the ALS-PoTQ method to the activations, we observe an accuracy degradation of approximately 1\%$\sim$3\% on different models. This is because of the PoT quantization's rigid resolution problem~\cite{li2019additive}. The rigid resolution problem means that the resolution in the long-tail area is sparse and fixed. As shown in Figure~\ref{fig:pot}, the 4-bit PoT quantization only has higher resolution than 3-bit PoT quantization in the small area close to zero. Since the format of PoT numbers cannot be changed, we can change the quantization range to improve the resolution of the long-tail area, which is the ``clipping'' technique in quantization works. 

Thus, we change the data range of $A$, inspired by PACT~\cite{choi2018pact}. The clipping threshold should be adaptive for different layers because the data distributions are different. To apply our technique to $A$ of all linear layers, we propose a Parameterized Ratio Clipping (PRC) technique with a clipping ratio factor $\gamma$. Given a set of FP32 activations $A = \{a_1, a_2, \cdots, a_n\}$ and a clipping ratio factor $\gamma$, the clipped activations $\bar{A}= \{ \bar{a_1}, \bar{a_2}, \cdots, \bar{a_n}\}$ is defined as:
\begin{equation}
\label{flip}
\bar{a_i}=\left\{
\begin{array}{lc}
-max(\vert A \vert)\cdot \gamma, \quad if \quad a_i < -max(\vert A \vert)\cdot \gamma,  \\ 
max(\vert A \vert)\cdot \gamma, \qquad if \quad a_i > max(\vert A \vert)\cdot \gamma, \\
a_i, \quad \qquad \qquad \qquad else.
\end{array}
\right.
\end{equation}

\begin{table}[t]
\centering
\caption{Energy consumption of different operations.}
\scalebox{0.95}{
\begin{tabular}{cccccc}
\toprule[1.2pt]
\multicolumn{6}{c}{Unit Energy Consumption(pJ)} \\ \hline 
\multirow{2}{*}{Multiplier} & FP32   & INT32   & FP8  & INT8  & INT4   \\ \cline{2-6}
                            & 3.7    & 3.1     & 0.23   & 0.19   & 0.048  \\ \hline 
\multirow{2}{*}{Adder} & FP32   & INT32   & INT16  & INT8  & INT4   \\ \cline{2-6}
                            & 0.9    & 0.14     & 0.05   & 0.03  & 0.015  \\ \hline
\multirow{2}{*}{Shift}    & INT32-4 & \multicolumn{2}{c}{INT32-3} & \multicolumn{2}{c}{INT4-3}   \\ \cline{2-6}
                         &  0.96 & \multicolumn{2}{c}{0.72} & \multicolumn{2}{c}{0.081}             \\  \bottomrule[1.2pt]
\end{tabular}
}
\label{tab:energy}
\end{table}
\section{Multiplication-Free Training Scheme}
\label{sec:mac}
In Section~\ref{sec:als-pot}, we propose the ALS-PoTQ method to obtain 5-bit PoT numbers. In this section, we further describe how we implement the Multiplication-Free MAC (MF-MAC) and ALS-PoTQ.

As shown in Figure~\ref{fig:compute_mac}, we replace the FP32 MAC (consisting of FP32 multiplication and  FP32 accumulation) with the ALS-PoTQ and MF-MAC. The inputs of MAC are two sets of FP32 numbers $\{x_{11}, x_{12}, \cdots, x_{ij}, \cdots, x_{mn}\}$ (shorten as $\{x_{ij}\}$, we call it a data block) and $\{x'_{ij}\}$. In our method, we use the ALS-PoTQ to convert both of the FP32 inputs first. In the ALS-PoTQ, the data block $\{x_{ij}\}$ (or $\{x'_{ij}\}$)  is first scaled by the layer-wise scaling factor $\alpha=2^{\beta}$, which is implemented as an INT8 addition between the integer $\beta$ and the exponent part of FP32 number to obtain scaled numbers $\{y_{ij}\}$. Then, we round the scaled numbers to the nearest PoT numbers $p=2^e$ to obtain an INT4 block $\{e_{ij}\}$ with a 1-bit sign block $\{s_{ij}\}$.

After the ALS-PoTQ, the FP32 data blocks $\{x_{ij}\}$ and $\{x_{ij}\}$ are converted to two INT4 data blocks $\{e_{ij}\}$ and $\{e'_{ij}\}$, two 1-bit sign data blocks $\{s_{ij}\}$ and $\{s'_{ij}\}$, and two integers $\beta$ and $\beta'$, which are the inputs of MF-MAC. In MF-MAC, we apply a INT4 adder to compute the INT4 addition between $\{p_{ij}\}$ and $\{p'_{ij}\}$. Meanwhile, we apply an XOR gate to process the sign flip operation in Equation~\eqref{eq:sign}. Then, we combine the output of the INT4 adder with the output of the XOR gate and accumulate the signed numbers to an INT32 number $z$. Finally, we shift the INT32 number $z$ by $\beta+\beta'$ bits to obtain the output of the MF-MAC.

Combining the ALS-PoTQ and MF-MAC with the proposed WBS and PRC techniques, we give a complete multiplication-free training scheme as shown in Algorithm~\ref{algor}. In the forward propagation, we first correct the bias of $W^l$ and clip $A^l$ to obtain $W_{unbias}^l$ and $A_{clipped}^l$. Then, 
we use the ALS-PoTQ method to convert $W_{unbias}^l$ and $A_{clipped}^l$ to 5-bit PoT numbers $W_q^l$ and $A_q^l$. After that, we apply the MF-MAC to obtain $A^{l+1}$ in the next layer. During backward propagation, we use the ALS-PoTQ to convert $G^l$ to 5-bit PoT numbers $G_q^l$ and apply the MF-MAC to obtain $G^{l-1}$ and $\Delta W^l$ to update $W^l$. We repeat the above processes until the network convergence.

\begin{algorithm}[t]
\caption{Forward and Backward Propagation in multiplication-free training.}
\label{algor}
\begin{algorithmic}[1]
\STATE \textbf{Definition:} $W^l$, $A^l$, and $G^l$ denote the full precision weights, activations, and gradients in layer $l$. $\gamma^l$ is the scaling factor of $A^l$. $L$ is the number of linear layers. $MF\_MAC()$ is the multiplication-free MAC operation.
\STATE \textbf{forward:}
\FOR{$l <= L$} 
\STATE $W_{unbias}^l = W^l - mean(W^l)$
\STATE $W_q^l = ALS\_PoT\_quantization(W_{unbias}^l)$
\STATE $A_{clipping}^l = clipping(A^l,\gamma^l)$
\STATE $A_q^l = ALS\_PoT\_quantization(A_{clipped}^l)$
\STATE $A^{l+1} = MF\_MAC\_pure(W_q^l, A_q^l)$
\STATE $l = l + 1$
\ENDFOR
\STATE \textbf{backward:}
\FOR{$l > 0$}
\STATE $G_q^l = ALS\_PoT\_quantization(G^l)$
\STATE $G^{l-1} = MF\_MAC\_pure(W_q^l, G_q^l)$
\STATE $\Delta W^{l-1} = MF\_MAC\_pure(A_q^l, G_q^l)$
\ENDFOR
\STATE \textbf{update}: $W^{1-L}$
\end{algorithmic}
\end{algorithm}

\begin{table*}[tbh]
\centering
\caption{``A'', ``W'', and ``G'' refer to activations, weights, and activation gradients. ``From Scratch'' refers to if the method trains the models from scratch or fine-tunes the pre-trained models. ``Multiplication'' refers to which operations are used to replace the multiplication in MAC. ``FW'' and ``BW'' refer to forward and backward propagation. ``Energy'' refers to the energy consumption of MACs for training ResNet50 on ImageNet at one iteration. ``*'' means ignoring the energy consumption of the multiplications in the quantization process. }
\scalebox{0.92}{
\begin{tabular}{ccccccccccc}
\toprule[1.2pt]
\multirow{3}{*}{Method} & \multirow{3}{*}{W} & \multirow{3}{*}{A} & \multirow{3}{*}{G} & \multicolumn{7}{c}{Training}                                                                                                                                                                                                                                                                                                                                                                                                                                           \\ \cline{5-11} 
                        &                    &                    &                    & \multirow{2}{*}{\begin{tabular}[c]{@{}c@{}}From\\  Scratch\end{tabular}} & \multirow{2}{*}{\begin{tabular}[c]{@{}c@{}}Large\\ Dataset\end{tabular}} & \multicolumn{2}{c}{Multiplication}                                                                                                      & \multicolumn{3}{c}{Energy (J)}                                                                                                                                         \\ \cline{7-11} 
                        &                    &                    &                    &                                                                          &                                                                          & FW                                                                 & BW                                                                 & FW                                                    & BW                                                    & Total                                                  \\ \hline
Original                & FP32               & FP32               & FP32               & -                                                                        & -                                                                        & FP32 Mul                                                           & FP32 Mul                                                           & 4.84                                                  & 9.69                                                  & 14.53                                                  \\ \hline
INQ                     & PoT5               & FP32               & FP32               & $\times$                                                                 & \checkmark                                                & \begin{tabular}[c]{@{}c@{}}FP32 Mul\\ (INT32-4 Shift)\end{tabular} & FP32 Mul                                                           & \begin{tabular}[c]{@{}c@{}}4.84\\ (1.97)\end{tabular} & 9.69                                                  & 14.53                                                  \\ \hline
LogNN                   & PoT4               & PoT4               & FP32               & $\times$                                                                 & $\times$                                                                 & \begin{tabular}[c]{@{}c@{}}FP32 Mul\\ (INT3 Add)\end{tabular}      & \begin{tabular}[c]{@{}c@{}}FP32 Mul\\ (INT32-3 Shift)\end{tabular} & \begin{tabular}[c]{@{}c@{}}4.84\\ (0.95)\end{tabular} & \begin{tabular}[c]{@{}c@{}}9.69\\ (1.92)\end{tabular} & \begin{tabular}[c]{@{}c@{}}14.53\\ (2.87)\end{tabular} \\ \hline
ShiftCNN                & PoT4               & FP32               & FP32               & $\times$                                                                 & \checkmark                                                & \begin{tabular}[c]{@{}c@{}}FP32 Mul\\ (INT32-4 Shift)\end{tabular} & FP32 Mul                                                           & \begin{tabular}[c]{@{}c@{}}4.84\\ (1.70)\end{tabular} & 9.69                                                  & 14.53                                                  \\ \hline
ShiftAddNet             & PoT5               & INT32              & INT32              & \checkmark                                                & $\times$                                                                 & \begin{tabular}[c]{@{}c@{}}INT32-4 Shift\\ INT32 Add\end{tabular}  & \begin{tabular}[c]{@{}c@{}}INT32 Mul\\ INT32-4 Shift\end{tabular}  & 2.45                                                  & 6.63                                                  & 9.08                                                   \\ \hline
AdderNet                & FP32               & FP32               & FP32               & \checkmark                                                & \checkmark                                                & FP32 Add                                                           & FP32 Add                                                           & 1.90                                                  & 3.80                                                  & 5.70                                                   \\ \hline
DeepShift-Q             & PoT5               & INT32              & FP32               & \checkmark                                                & \checkmark                                                & INT32-4 Shift                                                      & \begin{tabular}[c]{@{}c@{}}FP32 Mul\\ INT8 Add\end{tabular}        & 1.97                                                  & 5.84                                                  & 7.81                                                   \\ \hline
DeepShift-PS            & PoT5               & INT32              & FP32               & \checkmark                                                & \checkmark                                                & INT32-4 Shift                                                      & \begin{tabular}[c]{@{}c@{}}FP32 Mul\\ INT8 Add\end{tabular}        & 1.97                                                  & 5.84                                                  & 7.81                                                   \\ \hline
S2FP8              & FP8                & FP8               & FP8               & \checkmark                                                                 & \checkmark                                                & FP8 Mul & FP8 Mul                                                           & 1.19*  &      2.38*                                            & 3.57*                                                  \\ \hline
LUQ                     & INT4               & INT4               & PoT5               & \checkmark                                                & \checkmark                                                & INT4 Mul                                                           & Shift4-3                                                            & 1.00*                                                 & 2.06*                                                 & 3.07*                                                  \\ \hline
Ours                    & PoT5               & PoT5               & PoT5               & \checkmark                                                & \checkmark                                                & INT4 Add                                                           & INT4 Add                                                           & 0.16                                                  & 0.33                                                  & \textbf{0.49}                                          \\ \bottomrule[1.2pt]
\end{tabular}
}
\label{tab:total_compare}
\end{table*}

\vspace{-2mm}
\section{Energy Consumption Analysis}
\label{sec:energy}

In this section, we analyze the energy consumption of our method and compare it with the related works. First, Table~\ref{tab:energy} shows the unit energy consumption values of different operations implemented in 45nm CMOS technology, following ~\cite{wang2021addernet, you2020shiftaddnet}.

In our work, we replace each FP32 multiplication in the MAC with an INT4 addition and an XOR gate. As shown in Table~\ref{tab:energy}, the energy consumption of an INT4 addition is approximately only 0.4\% of the FP32 multiplication and the energy consumption of an XOR gate is less than 0.01 pJ~\cite{wang2021addernet}. In addition, we can replace the FP32 accumulator in MAC with an INT32 accumulator which can reduce 84\% of energy. Thus, our multiplication-free MAC can reduce approximately 96.6\% of energy compared with the FP32 MAC. Moreover, we take the extra energy consumption introduced by our PoT quantization into account. We introduce INT8 additions, rounding operations in the ALS-PoTQ, and a scalar INT32 bitwise shift after the accumulation in MF-MAC. These operations consume approximately 0.04 pJ for every number. A detailed analysis of energy consumption is in Appendix B. In summary, our multiplication-free MAC with PoT quantizer can reduce approximately 95.8\% of energy compared with the FP32 MAC. 

We take a comprehensive comparison with related works including the novel low-precision quantization method~\cite{cambier2020shifted} as well as the multiplication-free networks~\cite{zhou2017incremental, miyashita2016convolutional, gudovskiy2017shiftcnn, chen2020addernet, you2020shiftaddnet, elhoushi2021deepshift, chmiel2021logarithmic}. 
As shown in Table~\ref{tab:total_compare}, we list what operations each method uses to replace the FP32 multiplication in MAC during forward and backward propagation. According to the energy consumption of these operations in Table~\ref{tab:energy}, we compute the energy consumption of different methods for training ResNet50 on ImageNet at one iteration, whose details are described in Appendix C. These comparison results show that our method significantly outperforms the existing methods for the energy consumption of DNN training. 

In addition, there are some flaws in the existing methods while not in our method: INQ~\cite{zhou2017incremental}, LogNN~\cite{miyashita2016convolutional}, and ShiftCNN~\cite{gudovskiy2017shiftcnn} use FP32 pre-trained models instead of training from scratch, so they cannot reduce the energy consumption of training. LogNN~\cite{miyashita2016convolutional} and ShiftAddNet~\cite{you2020shiftaddnet} do not conduct experiments on large-scale datasets such as ImageNet. S2FP8~\cite{cambier2020shifted} and LUQ~\cite{chmiel2021logarithmic} introduce extra multiplications in the quantization process, which increase the energy consumption as stated in ~\cite{jin2022f8net}. 

\section{Experiments}

\subsection{Training Accuracy Results}  
\subsubsection{CNN models on ImageNet}
\label{sec:cnn}

We train official models provided by TensorFlow~\cite{abadi2016tensorflow} and TensorPack. The detailed hyperparameter settings are in Appendix D. It is important to note that the initializer of weight should be untruncated normal distribution instead of truncated normal distribution. For a comprehensive comparison, we evaluate our method for the image classification task, which is chosen by most quantization works to evaluate performance. We do experiments with AlexNet~\cite{krizhevsky2012imagenet}, ResNet18~\cite{he2016deep}, ResNet50~\cite{he2016deep} on the ILSVRC12 ImageNet classification dataset \cite{deng2009imagenet:}.

Table~\ref{tab:cnn} gives the comprehensive accuracy comparison with related works, including INQ~\cite{zhou2017incremental}, ShiftCNN~\cite{gudovskiy2017shiftcnn}, AdderNet~\cite{chen2020addernet}, DeepShift~\cite{elhoushi2021deepshift}, Ultra-low~\cite{sun2020ultra}, and LUQ~\cite{chmiel2021logarithmic}. Here, we do not compare with LogNN~\cite{miyashita2016convolutional} and ShiftAddNet~\cite{you2020shiftaddnet} because they do not apply their methods to the training on large-scale datasets such as ImageNet. We compare with the training-from-scratch results in DeepShift and LUQ, instead of their fine-tuning results. However, INQ and ShiftCNN start with the pre-trained FP32 models instead of training from scratch, so we show their inference accuracy results here.

\vspace{-3mm}
\subsubsection{Transformer model on WMT En-De task} 
\label{sec:transformer}

In addition, looking beyond CNNs, we apply our training scheme to the Transformer-base model~\cite{vaswani2017attention} on the WMT En-De dataset for machine translation. We do not change any hyperparameter for FP32 training and the official model provided by TensorFlow. 
Compared with Ultra-low~\cite{sun2020ultra} and LUQ~\cite{chmiel2021logarithmic}, we achieve the highest BLEU score, with less than 0.3\% BLEU score degradation as shown in Table~\ref{tab:transformer}.

\begin{table}[t]
\centering
\caption{CNN accuracy results on ImageNet. ``Bit-width'' refers to the bit-width to represent data. Accuracy refers to the accuracy results of different methods. $\Delta$ refers to the accuracy degradation compared with FP32 training. }
\scalebox{0.92}{
\begin{tabular}{ccccc}
\toprule[1.2pt]
Model                     & Method       & \begin{tabular}[c]{@{}c@{}}bit-width\\ W/A/G\end{tabular} & \begin{tabular}[c]{@{}c@{}}Accuracy\\ (\%)\end{tabular} & \begin{tabular}[c]{@{}c@{}}$\Delta$\\ (\%)\end{tabular} \\ \hline
\multirow{4}{*}{AlexNet}  & Original     & 32/32/32                                                  & 58.00                                                   & -                                                       \\ \cline{2-5} 
                          & INQ          & 5/32/32                                                   & 56.13                                                   & -1.87                                                   \\ \cline{2-5} 
                          & Ultra-low    & 4/4/4                                                     & 56.38                                                   & -1.62                                                   \\ \cline{2-5} 
                          & Ours         & 5/5/5                                                     & \textbf{57.22}                                                   & -0.78                                                    \\ \hline
\multirow{9}{*}{ResNet18} & Original     & 32/32/32                                                  & 70.10                                                   & -                                                       \\ \cline{2-5} 
                          & INQ          & 5/32/32                                                   & 68.98                                                   & -1.12                                                   \\ \cline{2-5} 
                          & ShiftCNN     & 4/32/32                                                   & 64.24                                                   & -5.86                                                   \\ \cline{2-5} 
                          & AdderNet     & 32/32/32                                                  & 67.00                                                   & -3.10                                                   \\ \cline{2-5} 
                          & DeepShift-Q  & 5/32/32                                                   & 65.32                                                   & -4.78                                                   \\ \cline{2-5} 
                          & DeepShift-PS & 5/32/32                                                   & 65.34                                                   & -4.76                                                   \\ \cline{2-5} 
                          & S2FP8    & 8/8/8                                                     &  \textbf{69.6}                                                   & -0.50                                                  \\ \cline{2-5} 
                          & Ultra-low    & 4/4/4                                                     & 68.27                                                   & -1.83                                                   \\ \cline{2-5} 
                          & LUQ          & 4/4/4                                                     & 69.0                                                   & -1.10                                                   \\ \cline{2-5} 
                          & Ours         & 5/5/5                                                     & 69.52                                                   & -0.58                                                   \\ \hline
\multirow{9}{*}{ResNet50} & Original     & 32/32/32                                                  & 76.32                                                   & -                                                       \\ \cline{2-5} 
                          & INQ          & 5/32/32                                                   & 74.81                                                   & -1.51                                                   \\ \cline{2-5} 
                          & ShiftCNN     & 4/32/32                                                   & 72.58                                                   & -3.74                                                   \\ \cline{2-5} 
                          & AdderNet     & 32/32/32                                                  & 74.9                                                    & -1.42                                                   \\ \cline{2-5} 
                          & DeepShift-Q  & 5/32/32                                                   & 70.73                                                   & -5.59                                                   \\ \cline{2-5} 
                          & DeepShift-PS & 5/32/32                                                   & 71.90                                                   & -4.42                                                   \\ \cline{2-5} 
                          & S2FP8    & 8/8/8                                                     &  75.2                                                   & -1.12                                                  \\ \cline{2-5}
                          & Ultra-low    & 4/4/4                                                     & 74.01                                                   & -2.31                                                   \\ \cline{2-5} 
                          & LUQ          & 4/4/4                                                     & 75.32                                                   & -1.00                                                    \\ \cline{2-5} 
                          & Ours         & 5/5/5                                                     & \textbf{75.36}                                                   & -0.96                                                        \\ \bottomrule[1.2pt]
\end{tabular}
}
\label{tab:cnn}
\end{table}

\begin{table}[t]
\caption{BLEU results on WMT En-De tasks. ``bit-width'' refers to the bit-width to represent data.  $\Delta$ refers to the BLEU degradation compared with FP32 training.}
\centering
\scalebox{1.0}{
\begin{tabular}{ccccc}
\toprule[1.2pt]
Model                                                                        & Method    & \begin{tabular}[c]{@{}c@{}}bit-width\\ W/A/G\end{tabular} & \begin{tabular}[c]{@{}c@{}}BLEU\\ (\%)\end{tabular} & \begin{tabular}[c]{@{}c@{}}$\Delta$\\ (\%)\end{tabular} \\ \hline
\multirow{4}{*}{\begin{tabular}[c]{@{}c@{}}Transformer\\ -base\end{tabular}} & Original  & 32/32/32                                                  & 27.5                                                & -                                                     \\ \cline{2-5} 
                                                                             & Ultra-low & 4/4/4                                                     & 25.4                                                & -2.1                                                    \\ \cline{2-5} 
                                                                             & LUQ       & 4/4/4                                                     & \textbf{27.2}                                                & -0.3                                                    \\ \cline{2-5} 
                                                                             & Ours      & 5/5/5                                                     & \textbf{27.2}                                                & -0.3                                                    \\ \bottomrule[1.2pt]
\end{tabular}
}
\label{tab:transformer}
\end{table}

\subsection{Accuracy-Energy Joint Comparison}
\label{sec:pareto}

Synthesizing the previous experimental results, we give an energy-accuracy joint comparison whose result is shown in Fig.~\ref{fig:pareto}. The x-axis refers to the accuracy results and the y-axis refers to the energy consumption of both forward and backward propagation at one iteration. The joint comparison shows that our method has the lowest energy consumption as well as the highest accuracy among the methods that try to reduce the energy consumption of training.

\subsection{Ablation Study}
\label{sec:ablation}
In this section, we conduct an ablation study for the proposed techniques, including Adaptive Layer-wise Scaling PoT Quantization, Weight Bias Correction, and Parameterized Ratio Clipping.
The training accuracy results of ResNet18 in Table~\ref{Tab:ablation} proves the effects of these techniques. If there is no layer-wise scaling, the training collapses and accuracy drops to 0\%. This is because the representation range of PoT quantization cannot accommodate the data range, especially for the gradients. If there is no weight bias correction, the training is unstable, which is consistent with the analysis in Section~\ref{sec:weight_correct}. Moreover, the Parameterized Ratio Clipping technique improves the accuracy by 1.3\% for ResNet50. 

\begin{table}[t]
\centering
\caption{Comparison of Adaptive Layer-wise PoT Scaling (ALPS) , Parameterized Ratio Clipping (PRC) and Weight Bias Correction (WBC) for ResNet-50 on ImageNet.}
\begin{tabular}{cccccc}
\toprule[1.2pt]
ALS      & $\times$ & \checkmark & \checkmark &\checkmark & \checkmark \\ \hline
WBC         & $\times$ & $\times$   & \checkmark & $\times$ & \checkmark \\ \hline
PRC
 & $\times$ & $\times$      & $\times$   &\checkmark & \checkmark \\ \hline 
Accuracy(\%)              & 0.0     &  12.0/74.2   &    74.1   & 13.6 & 75.4\\ \bottomrule[1.2pt]
\end{tabular}
\label{Tab:ablation}
\end{table}

\section{Conclusion}
In this paper, we propose an Adaptive Layer-wise Scaling PoT Quantization (ALS-POTQ) method and a Multiplication-Free MAC (MF-MAC) to replace the FP32 multiplication in the original MAC with an INT4 addition and an XOR operation. We reduce up to 95.8\% of energy consumption in linear layers during training, with an accuracy degradation of less than 1\%. In summary, we significantly outperform the existing methods for both energy efficiency and accuracy.

{\small
\bibliographystyle{ieee_fullname}
\bibliography{main}
}

\clearpage

\setcounter{figure}{5}
\setcounter{table}{5}
\section*{APPENDIX}
\begin{figure}[t!]
  \centering
  \begin{subfigure}{0.49\linewidth}
    \includegraphics[scale=0.25]{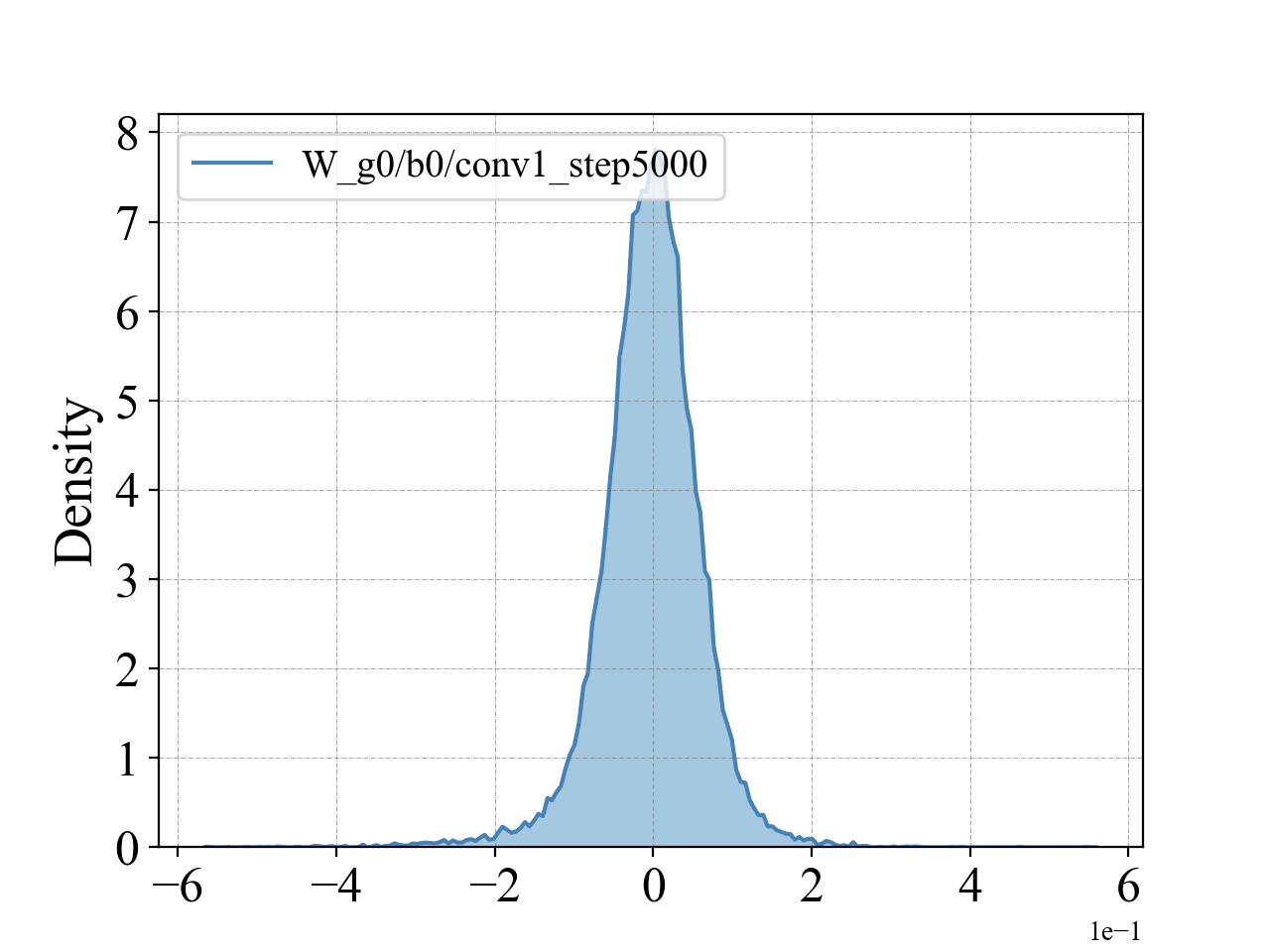}
    \caption{ResNet50 $W$}
  \end{subfigure}
  \hfill
    \begin{subfigure}{0.49\linewidth}
    \includegraphics[scale=0.25]{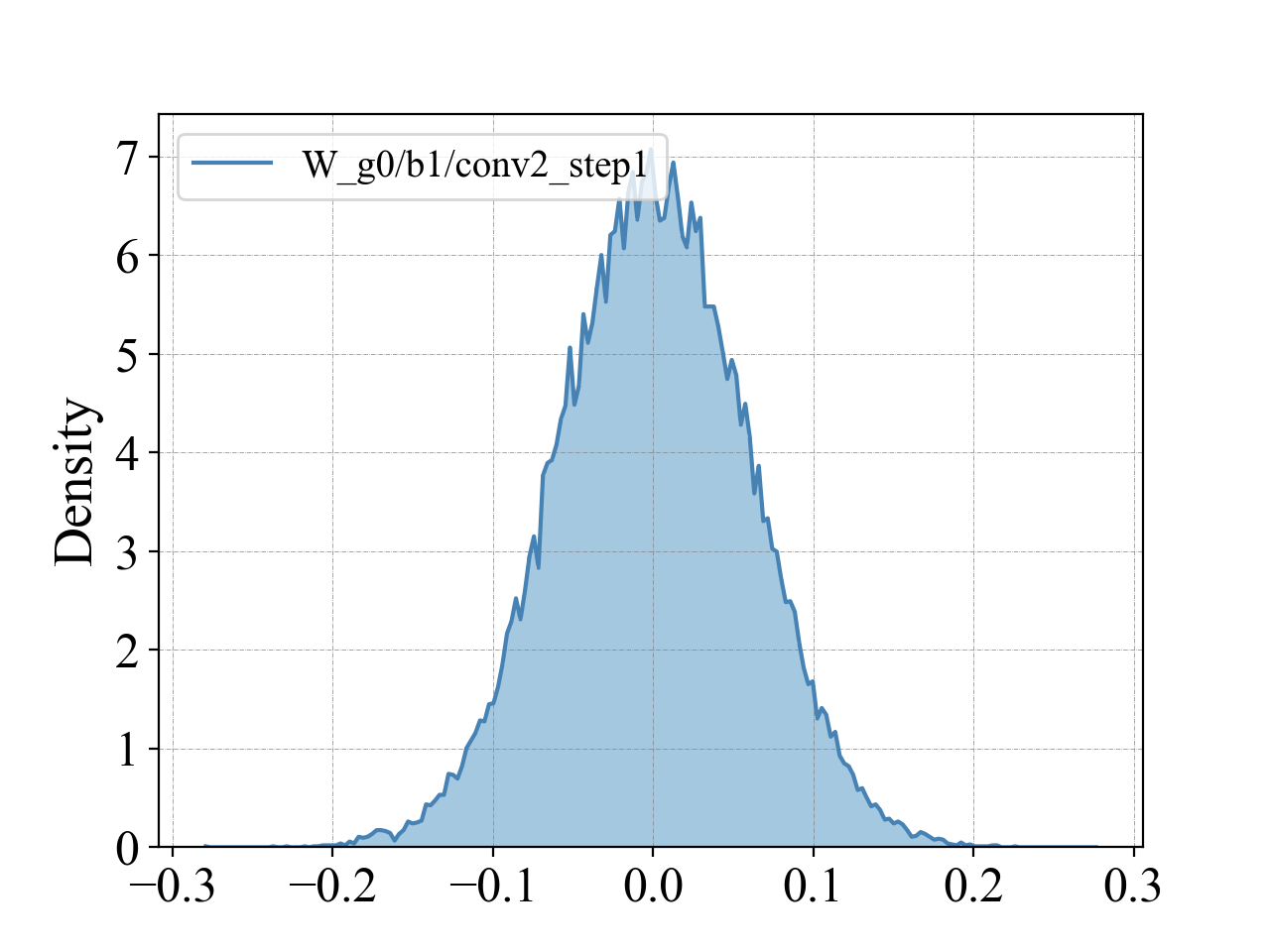}
    \caption{ResNet18 $W$}
  \end{subfigure}
  \hfill
  \begin{subfigure}{0.49\linewidth}
    \includegraphics[scale=0.25]{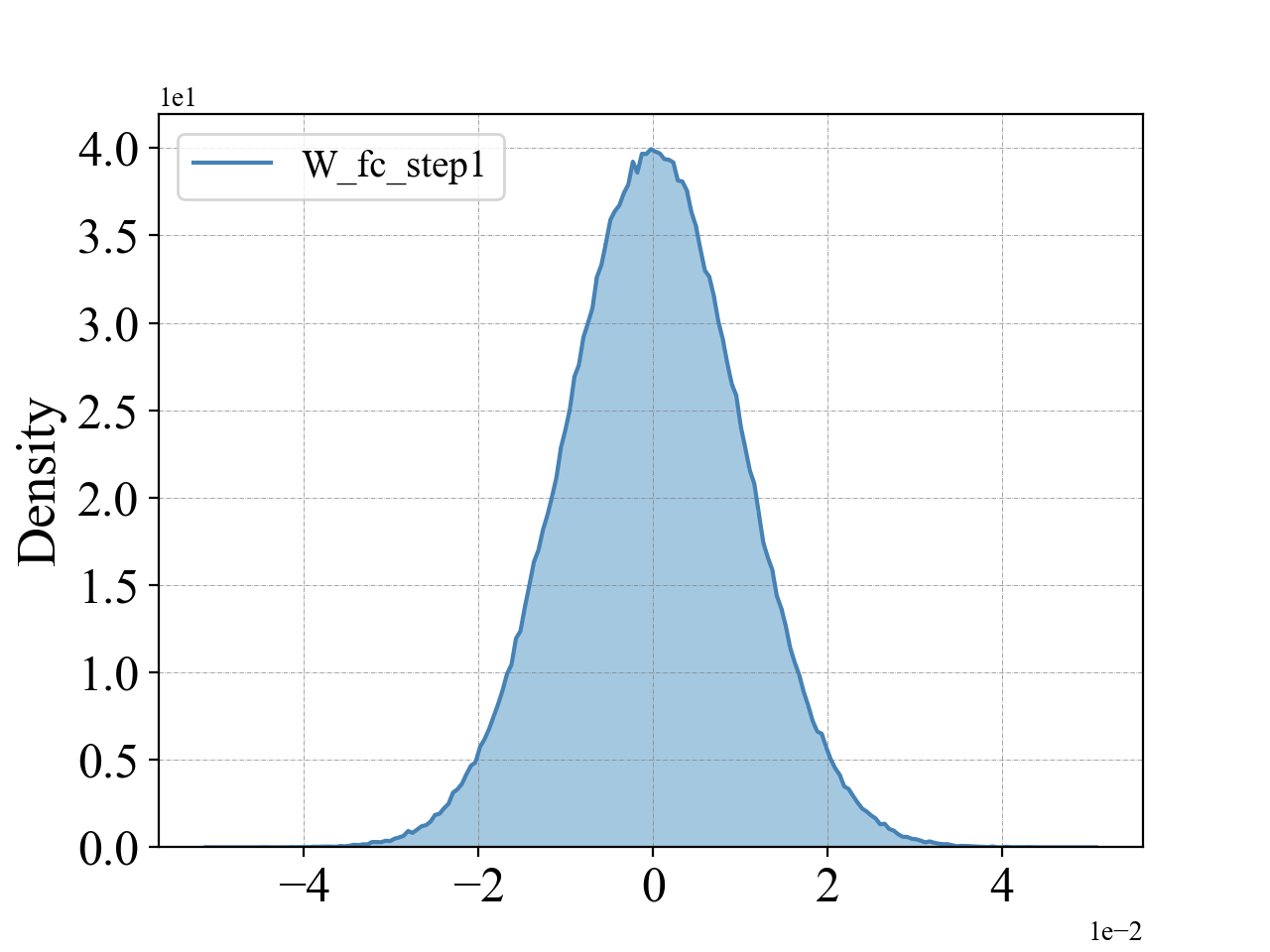}
    \caption{ResNet18 $W$}
  \end{subfigure}
  \begin{subfigure}{0.49\linewidth}
    \includegraphics[scale=0.25]{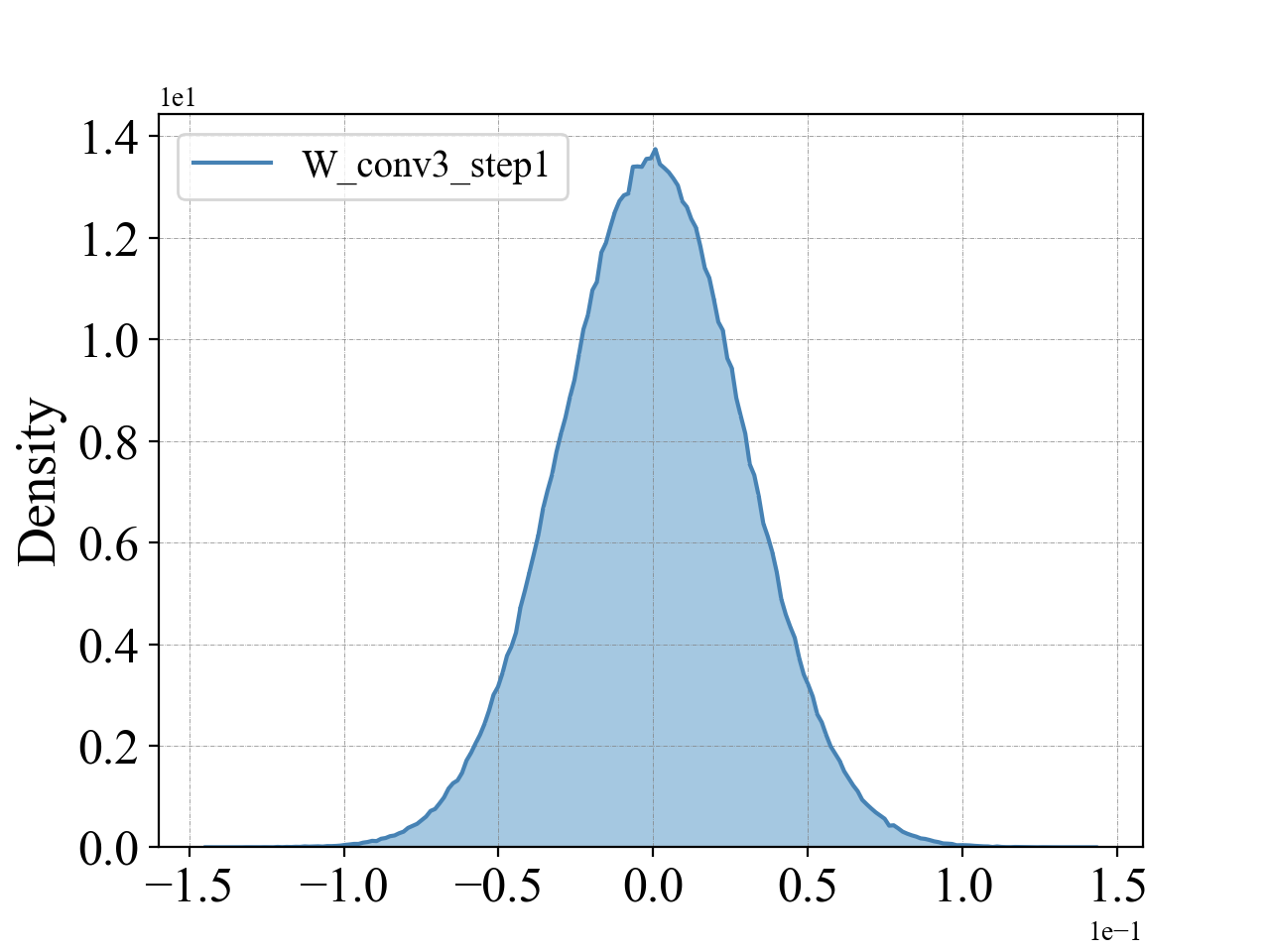} 
    \caption{AlexNet $W$}
  \end{subfigure}
  \hfill
    \begin{subfigure}{0.49\linewidth}
    \includegraphics[scale=0.25]{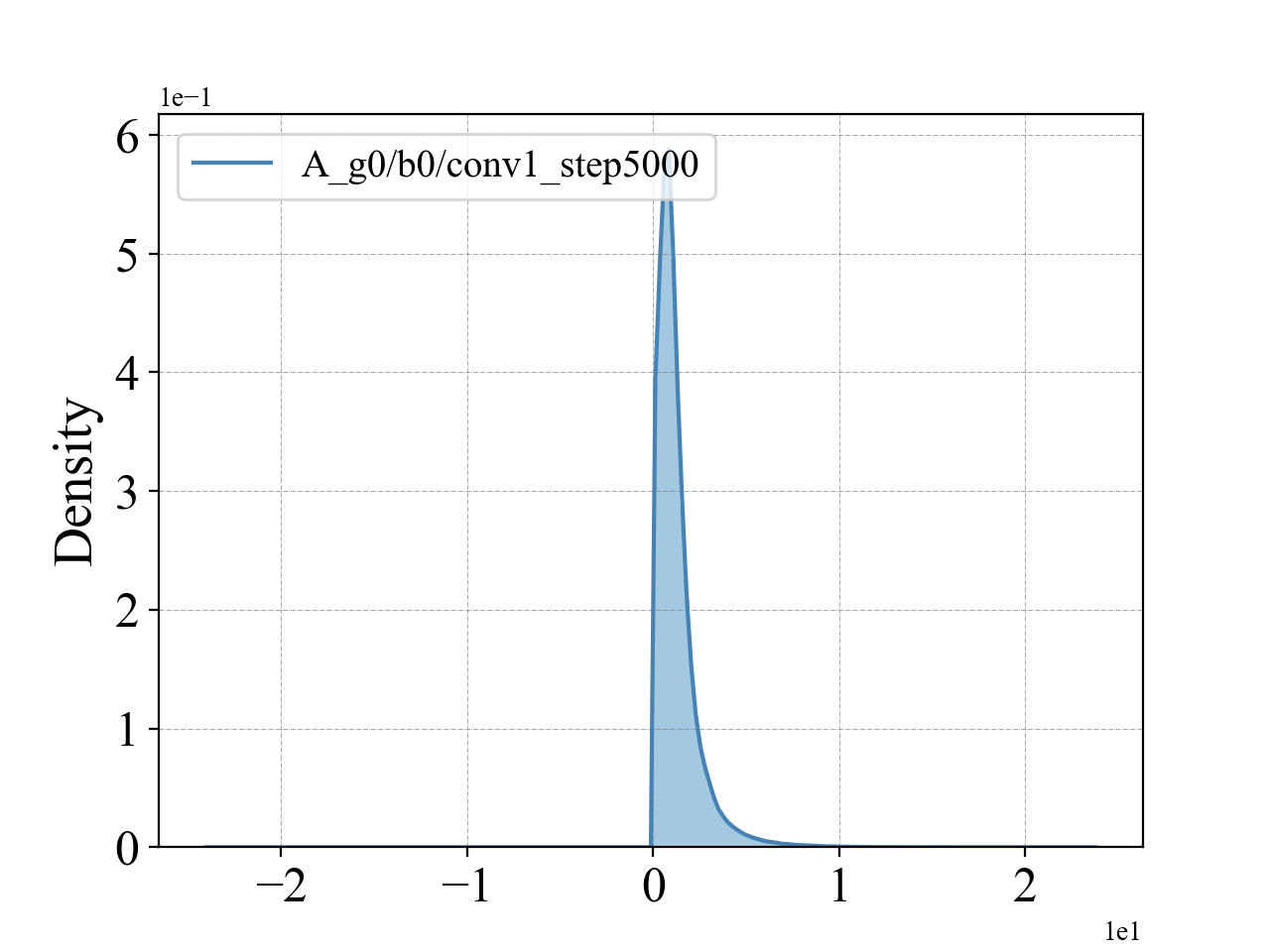}
    \caption{ResNet50 $A$}
  \end{subfigure}
  \hfill
  \begin{subfigure}{0.49\linewidth}
    \includegraphics[scale=0.25]{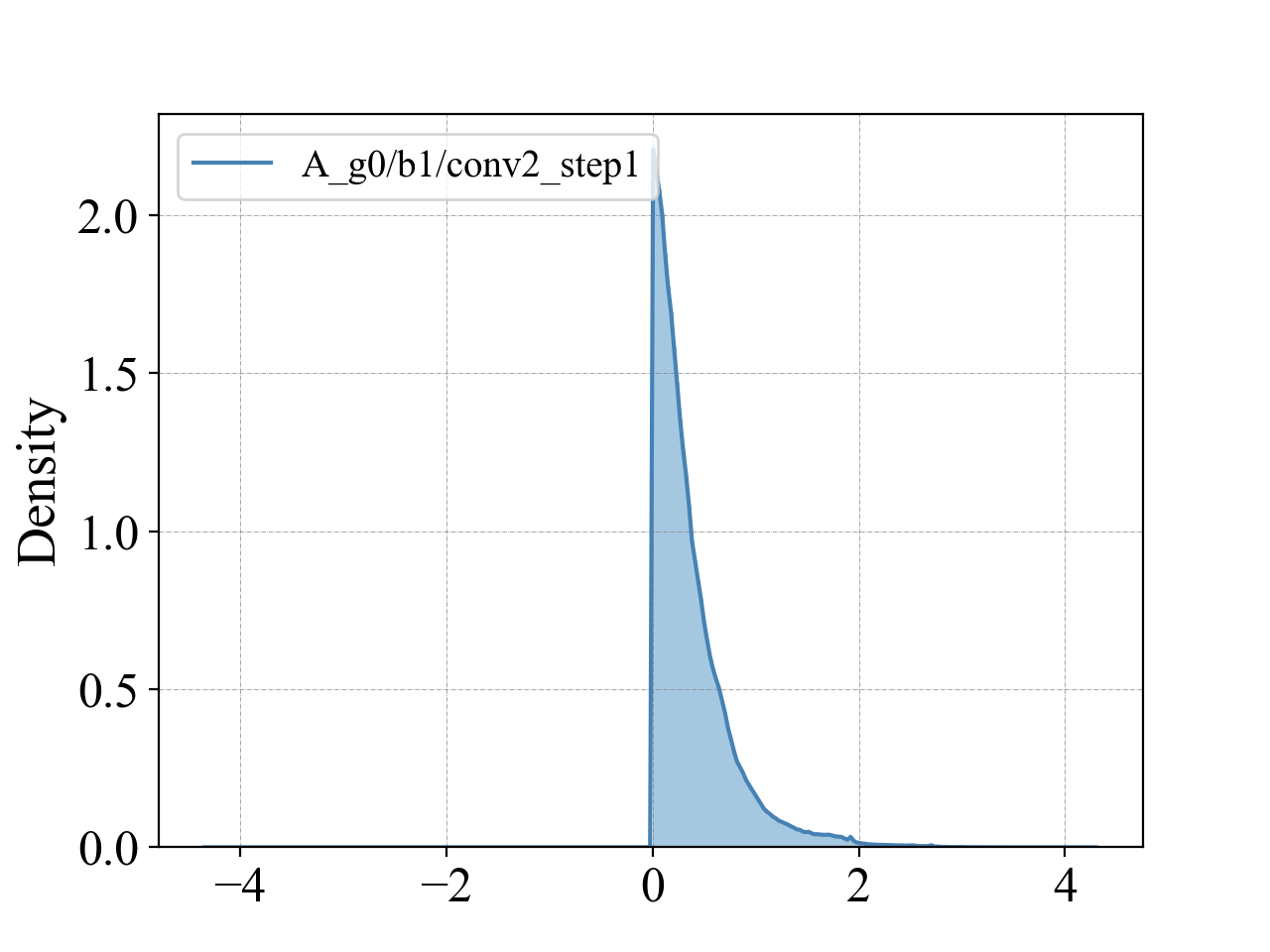}
    \caption{ResNet18 $A$}
  \end{subfigure}
   \hfill 
    \begin{subfigure}{0.49\linewidth}
    \includegraphics[scale=0.25]{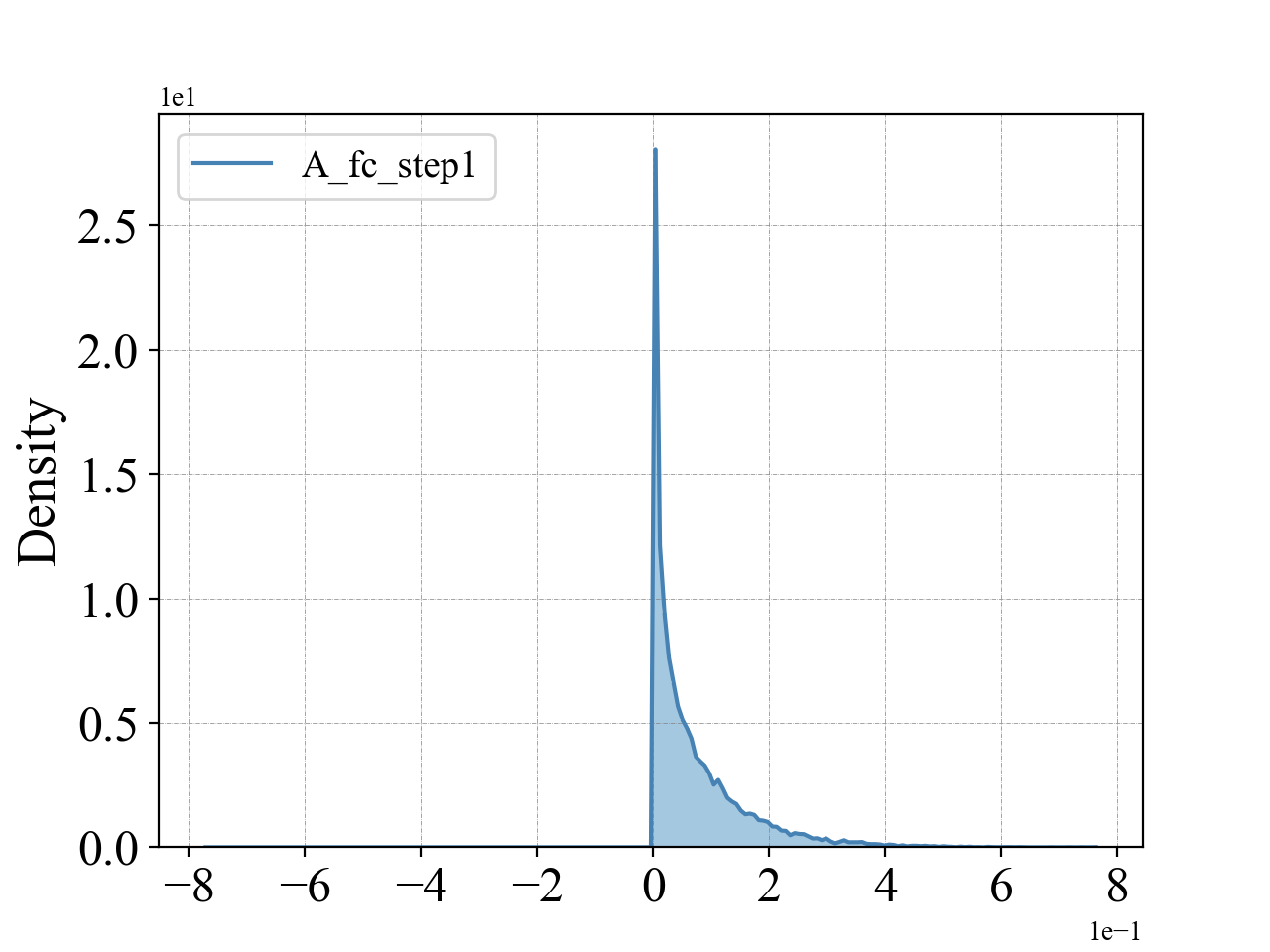}
    \caption{ResNet18 $A$}
  \end{subfigure}
  \hfill
    \begin{subfigure}{0.49\linewidth}
    \includegraphics[scale=0.25]{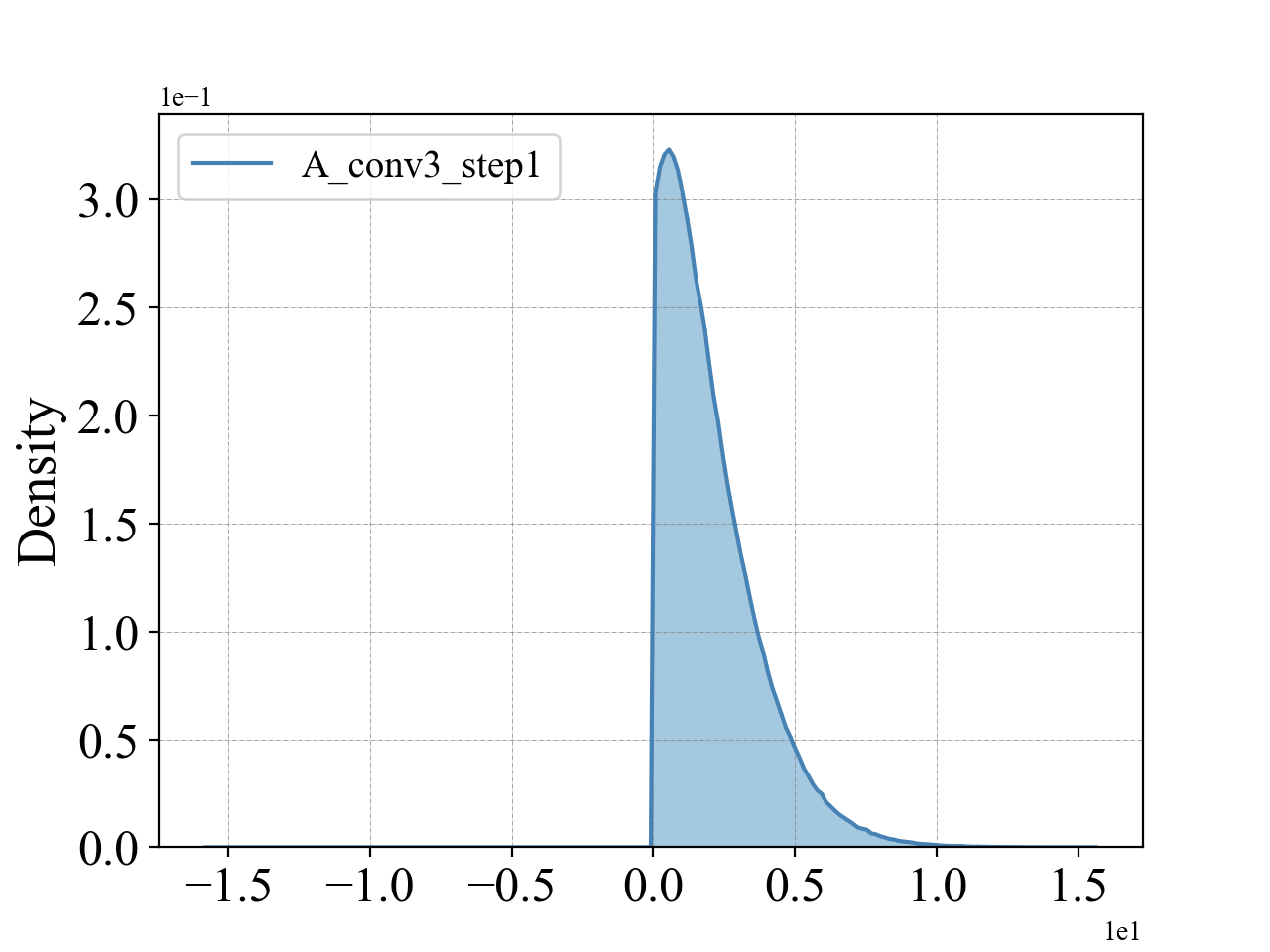} %
    \caption{AlexNet $A$}
  \end{subfigure}
  \hfill
  \begin{subfigure}{0.49\linewidth}
    \includegraphics[scale=0.25]{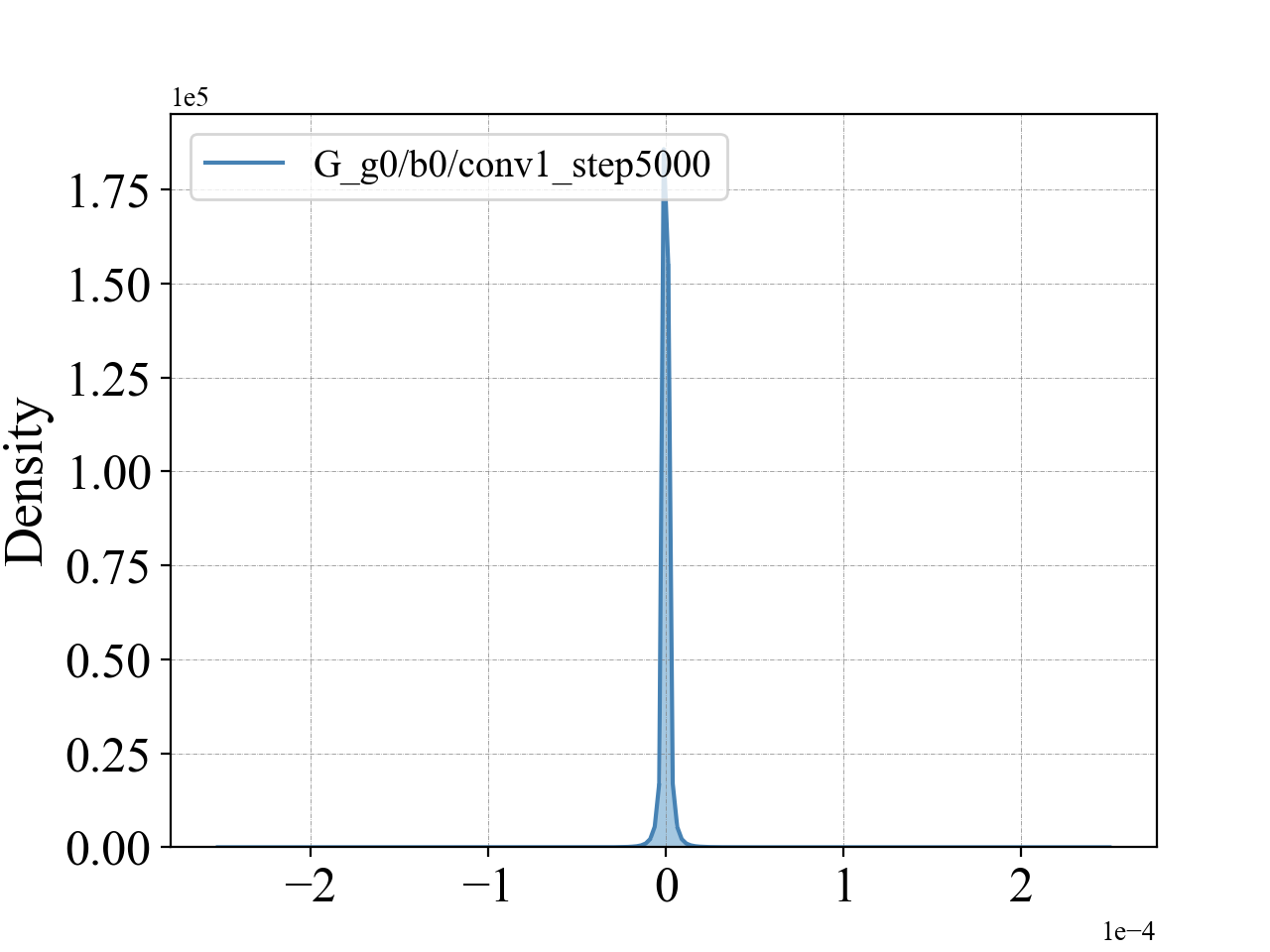}
    \caption{ResNet50 $G$}
  \end{subfigure}
  \begin{subfigure}{0.49\linewidth}
    \includegraphics[scale=0.25]{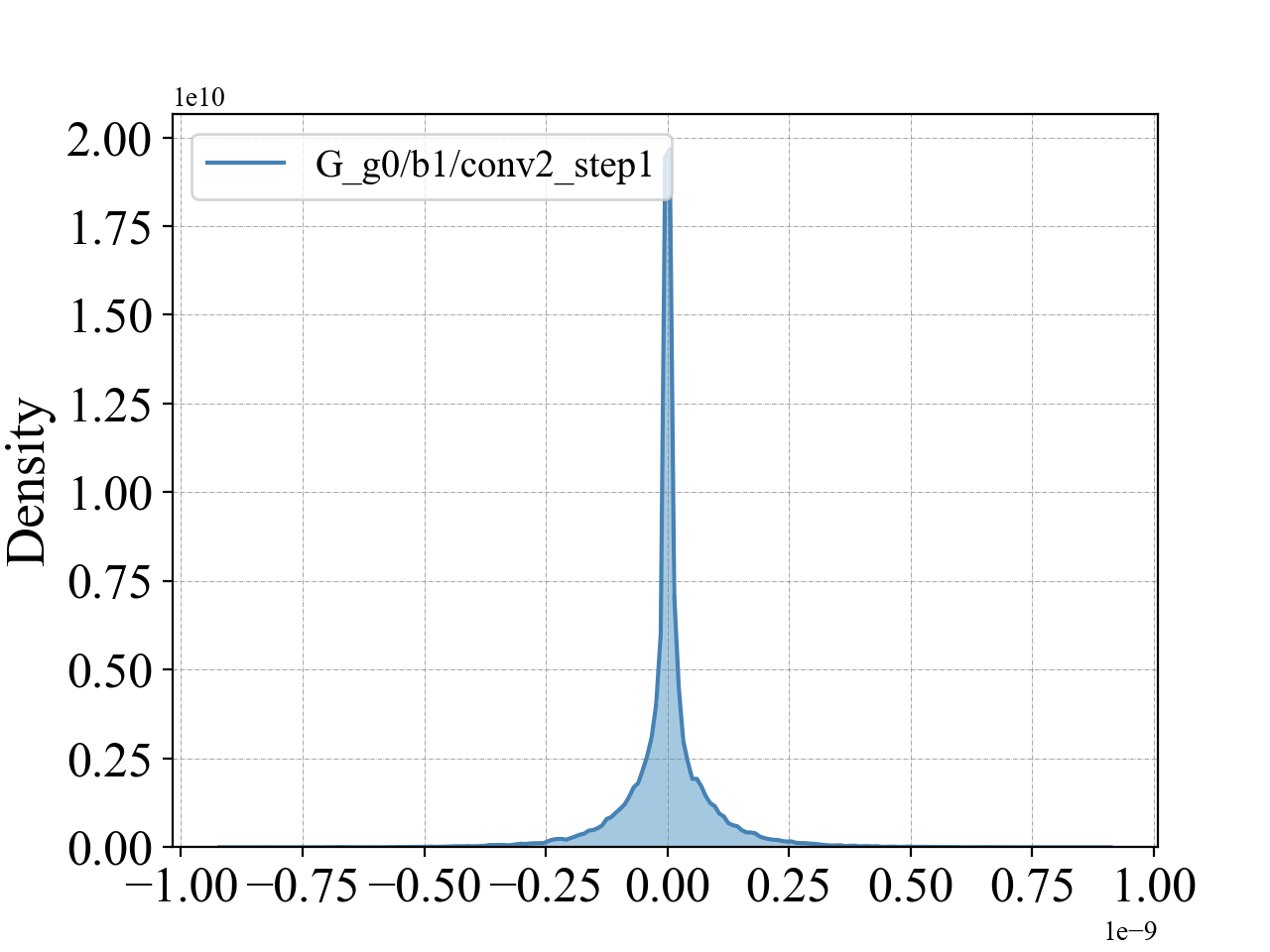}
    \caption{ResNet18 $G$}
  \end{subfigure}  \hfill
    \begin{subfigure}{0.49\linewidth}
    \includegraphics[scale=0.25]{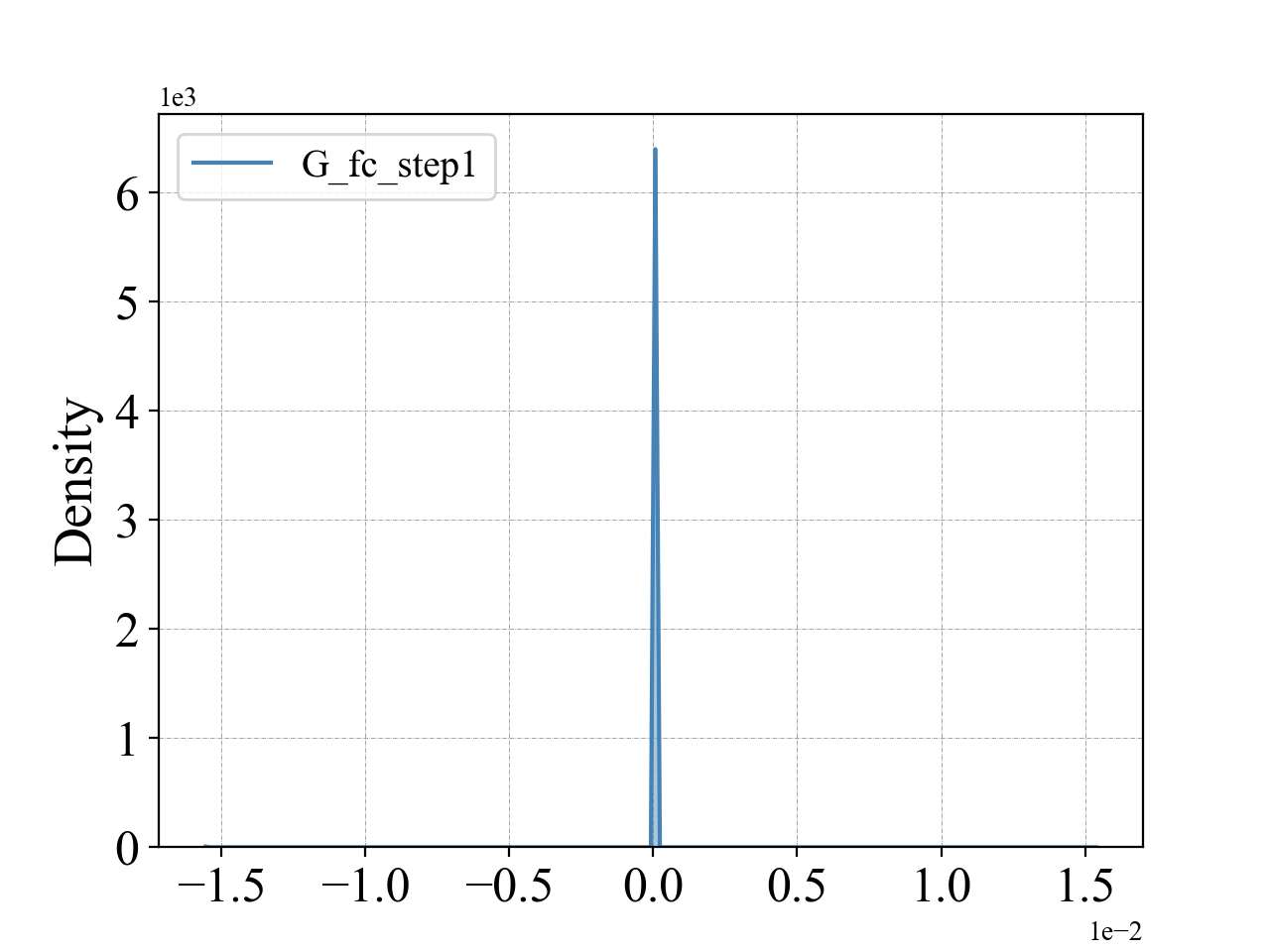}
    \caption{ResNet18 $G$}
  \end{subfigure}
  \hfill
  \begin{subfigure}{0.49\linewidth}
    \includegraphics[scale=0.25]{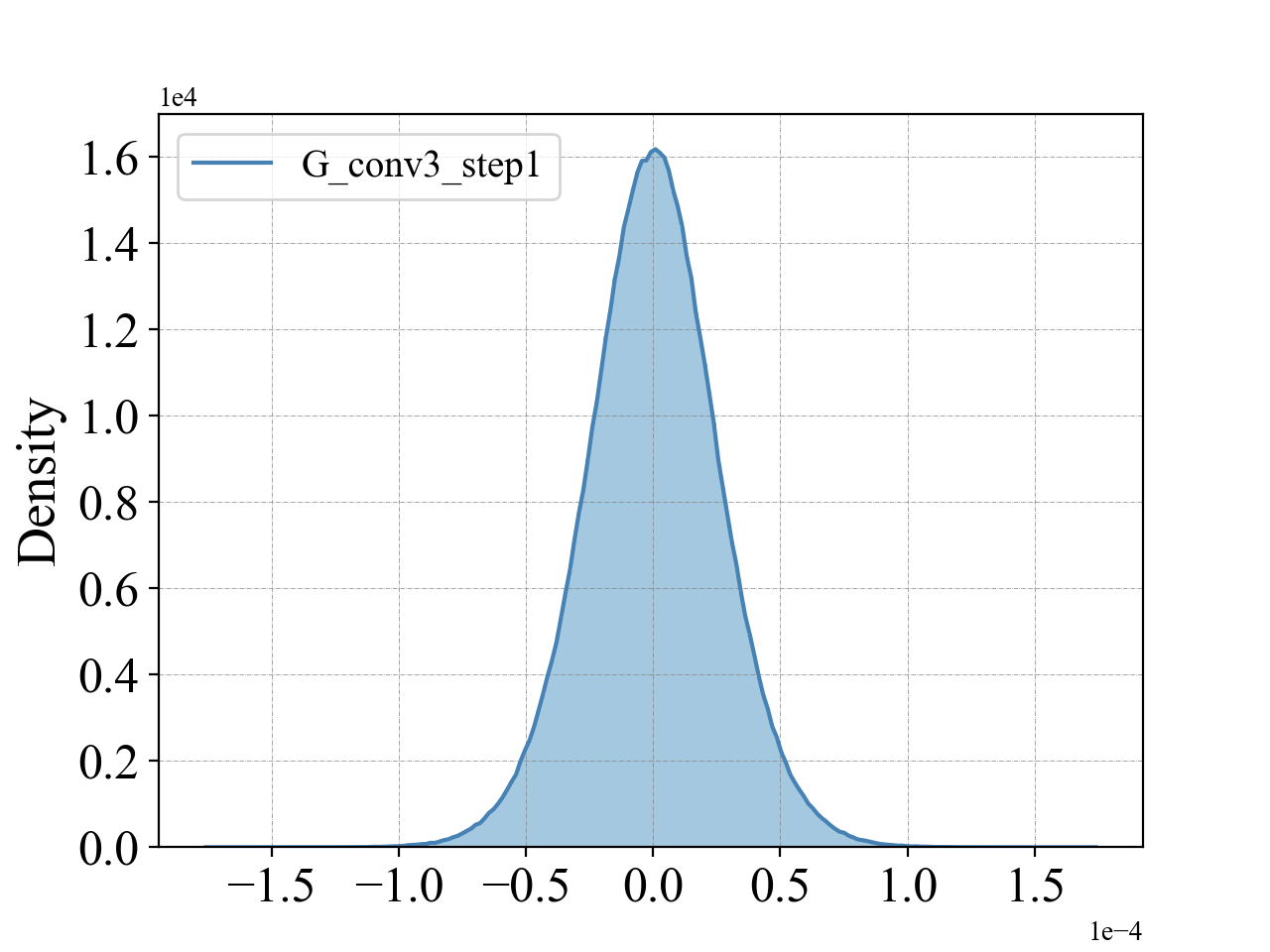}%
    \caption{AlexNet $G$}
  \end{subfigure}
  \caption{Distributions of $W$, $A$, and $G$. }
  \label{fig:more_distribution}
\end{figure}

\renewcommand\thesection{\Alph{section}}
\setcounter{section}{0}
\section{Data Distribution}
\label{sec:dist}

As shown in Figure~\ref{fig:more_distribution}, we give more distributions of $W$, $A$, and $G$ to support the observation in Section 4.1. 

\section{Energy Consumption of ALS-PoTQ}

We give a detailed analysis of the proposed ALS-PoTQ method's energy consumption.

In the proposed ALS-PoTQ method, we use an INT8 addition to scale each item in data block $x_{ij}$ whose size is $m \cdot n$, and thus the energy consumption of scaling is $(0.03\cdot m \cdot n) pJ$. Then, we round the scaled FP32 data block $y_{ij}$ to PoT numbers. After scaling, the exponent part of FP32 number $y_{ij}$ only takes 4-bit actually, and thus the round operation is a carry operation for INT4, which requires 4 half adders. However, the probability of carry operation is 50\%, that is, the round operation can be bypassed with a 50\% probability, then the dynamic power consumption of the round operation is approximately 1/4 of the 4-bit addition, which requires 3 full adders and 1 half adder. Thus, the energy consumption of a round operation is approximately $0.004 pJ$ and the total energy consumption of rounding the data block is $(0.004\cdot m \cdot n) pJ$. In summary, the energy consumption of ALS-PoTQ is $(0.034\cdot m \cdot n) pJ$. 

In addition, after the MF-MAC, we apply an INT32 shift to conduct the dequantization process. Because there is only one INT32 shift for the data blocks whose size is $m \cdot n$, the energy consumption of the INT32 shift is less than 0.05 for each number. In our training scheme, we apply both the ALS-PoTQ and MF-MAC three times during forward and backward propagation, so our ALS-PoTQ method consumes approximately $0.04 pJ$ for each number on average. Thus, the total energy consumption of an ALS-PoTQ and a MF-MAC is approximately $0.195pJ$.


\section{Energy Consumption of Existing Methods}

We compute the energy consumption of existing methods, including INQ~\cite{zhou2017incremental}, LogNN~\cite{miyashita2016convolutional}, ShiftCNN~\cite{gudovskiy2017shiftcnn},  AdderNet~\cite{chen2020addernet}, DeepShift~\cite{elhoushi2021deepshift},  ShiftAddNet~\cite{you2020shiftaddnet}, S2FP8~\cite{cambier2020shifted}, and LUQ~\cite{chmiel2021logarithmic}. First, we give which operations are used in MACs for forward and backward propagation. Then, we compute the energy consumption of MACs for training ResNet50 on ImageNet with the data in Table 1. There are $12.36 G$ MACs for training ResNet50 on ImageNet at one iteration. We compute the total energy consumption by multiplying the energy consumption of operations in a MAC with the MAC numbers. 

INQ~\cite{zhou2017incremental} and ShiftCNN~\cite{gudovskiy2017shiftcnn} convert $W$ to 5-bit or 4-bit PoT by fine-tuning the pre-trained models. Thus, the MACs during forward and backward propagation consist of FP32 multiplications and FP32 additions for training, while the FP32 multiplications in MACs during forward for inference are replaced with INT32-4 bitwise shift (shifting up to 4-bit on INT32 numbers).
LogNN~\cite{miyashita2016convolutional} converts $W$ and $A$ to 4-bit PoT by fine-tuning the pre-trained models, which is similar or INQ. In addition, it also try to train from scratch, however, it does not conduct experiments on large-scale datasets such as ImageNet. 
AdderNet~\cite{chen2020addernet} replaces the FP32 multiplications in MACs with FP32 additions, so there are two FP32 additions in a MAC operation. 
DeepShift~\cite{elhoushi2021deepshift} converts $W$ to 5-bit PoT numbers and trains the models from scratch. Thus, the multiplications during forward propagation are replaced with INT32-4 bitwise Shift, and half of the multiplications ($WG$) during backward propagation can be replaced with INT8 additions on the exponent part of FP32 $G$.
ShiftAddNet~\cite{you2020shiftaddnet} combines the bitwise shift operations in DeepShift and the additions in AdderNet, so it replaces the multiplications during forward propagation to INT32-4 bitwise shifts and FP32 additions, and replaces half of the multiplications during backward propagation to INT32-4 bitwise shifts.

Moreover, S2FP8~\cite{cambier2020shifted} quantizes $W$, $A$, and $G$ to FP8 numbers, and the multiplications in MAC for forward and backward propagation are replaced with FP8 multiplications. Ultra-low~\cite{sun2020ultra} uses radix-4 float point numbers, which are not supported by the radix-2 hardware, so we do not compute its energy consumption here. LUQ~\cite{chmiel2021logarithmic} quantizes $W$ and $A$ to INT4 numbers, and converts $G$ to 4-bit PoT numbers. Thus, it replaces the multiplications during forward propagation to INT4 multiplications and the multiplications during backward propagation to INT4-3 bitwise shifts. In addition, these three quantization training works introduce extra FP32 multiplications in their methods. 
To avoid ambiguity, we do not compute the energy consumption of these extra multiplications and ignore these multiplications when computing their energy consumption.

In total, we compute the MAC energy consumption in these works based on the energy consumption data in Table 1 and the analyses of operations.

\section{Training Settings}

We train AlexNet~\cite{krizhevsky2012imagenet} on 4 GPUs with standard hyperparameter: 100 epochs, batch size 256, SGD with the momentum of 0.9, the initial learning rate of 0.02 decreased by a factor of 10 after epoch 30, 60, 90; ResNet18 and ResNet50~\cite{he2016deep} on 8 GPUs with standard hyperparameter: 105 epochs, batch size of 256, SGD with the momentum of 0.9, the initial learning rate of 0.1 decreased by a factor of 10 after epoch 30, 60, 90. In addition, all of the weights are initialized as untruncated normal distributions and we convert $G$ in the last layer to 6-bit PoT numbers instead of 5-bit PoT numbers.

\section{Accuracy result on ResNet101}
We apply our method to deeper network ResNet101 and also achieve high accuracy as shown in Table~\ref{tab:resnet101}. 
\begin{table}[t]
\centering
\caption{CNN accuracy results on ImageNet. ``Bit-width'' refers to the bit-width to represent data. Accuracy refers to the accuracy results of different methods. $\Delta$ refers to the accuracy degradation compared with FP32 training. }
\scalebox{0.92}{
\begin{tabular}{ccccc}
\toprule[1.2pt]
Model                     & Method       & \begin{tabular}[c]{@{}c@{}}bit-width\\ W/A/G\end{tabular} & \begin{tabular}[c]{@{}c@{}}Accuracy\\ (\%)\end{tabular} & \begin{tabular}[c]{@{}c@{}}$\Delta$\\ (\%)\end{tabular} \\ \hline
\multirow{2}{*}{ResNet101}  & Original     & 32/32/32                                                  & 78.05                                                   & -                                                       \\ \cline{2-5} 
                          & Ours         & 5/5/5                                                     & \textbf{77.21}                                                   & -0.84                                                    \\  \bottomrule[1.2pt]
\end{tabular}
}
\label{tab:resnet101}
\end{table}

\end{document}